%% file: main.tex
\documentclass{article} 
\usepackage{arxiv}  
\usepackage[utf8]{inputenc} 
\usepackage[T1]{fontenc}    
\usepackage{hyperref}       
\usepackage{url}            
\usepackage{booktabs}       
\usepackage{amsfonts}       
\usepackage{nicefrac}       
\usepackage{microtype}      
\usepackage{lipsum}

\usepackage{times}
\usepackage{soul}
\usepackage{url}
\usepackage[small]{caption}
\usepackage{graphicx}
\usepackage{amsmath}
\usepackage{amsthm}
\usepackage{booktabs}
\usepackage{algorithm}
\usepackage{todonotes}
\usepackage{mathtools}
\usepackage{multirow}

\usepackage{amssymb}
\usepackage{natbib}
\usepackage{hyperref}

\input{pckgs}

\input{defs}

\usepackage{newfloat}
\usepackage{listings}
\DeclareCaptionStyle{ruled}{labelfont=normalfont,labelsep=colon,strut=off} 
\floatstyle{ruled}
\newfloat{listing}{tb}{lst}{}
\floatname{listing}{Listing}
\title{Sequential Information Design:\\Learning to Persuade in the Dark}
\author{
	Martino Bernasconi\\
	Politecnico di Milano\\
	\texttt{martino.bernasconideluca@polimi.it}
	\And
	Matteo Castiglioni\\
	Politecnico di Milano\\
	\texttt{matteo.castiglioni@polimi.it}
	\And
	Alberto Marchesi\\
	Politecnico di Milano\\
	\texttt{alberto.marchesi@polimi.it}
	\And
	Nicola Gatti\\
	Politecnico di Milano\\
	\texttt{nicola.gatti@polimi.it}
	\And
	Francesco Trovò\\
	Politecnico di Milano\\
	\texttt{francesco1.trovo@polimi.it}
}

\begin{document}

\maketitle

\input{src/abstract}
\input{src/introduction}
\input{src/related_works}
\input{src/preliminaries}
\input{src/online_problem}

\input{src/characterization}
\input{src/offline}

\input{src/online_learning}
\input{src/full_feedback}
\input{src/bandit_feedback}
\input{src/bandit_due}

\clearpage

\bibliographystyle{plainnat}
\bibliography{biblio}

\clearpage

\appendix
\input{src/appendix}

\end{document}

%% file: pckgs.tex
\usepackage{amsfonts}
\usepackage{multicol}
\usepackage{algorithm}
\usepackage[noend]{algpseudocode}
\usepackage{mathtools}
\usepackage{amsthm}
\usepackage{thmtools}
\usepackage{thm-restate}
\usepackage{lipsum}
\usepackage{amssymb }
\usepackage{mleftright}
\usepackage{color}
\usepackage{diagbox}
\usepackage{tikz}
\usepackage{wrapfig}
\usetikzlibrary{positioning,fit,calc,matrix,arrows,trees,backgrounds}
\usepackage{accents}
\usepackage{tikz}
\usepackage{pgfplots}
\usepackage{enumerate}
\usepackage[shortlabels]{enumitem}
\usepackage[utf8]{inputenc} 
\usepackage[T1]{fontenc}    
\usepackage{hyperref}       
\usepackage{url}            
\usepackage{booktabs}       
\usepackage{amsfonts}       
\usepackage{nicefrac}       
\usepackage{microtype}      
\usepackage{xcolor}         

%% file: defs.tex
\newcommand{\I}{\mathcal{I}}

\newcommand{\A}{\mathcal{A}}

\newcommand{\Z}{\mathcal{Z}}

\newcommand{\C}{\mathcal{C}}
\newcommand{\B}{\mathcal{B}}
\newcommand{\Hi}{\mathcal{H}}

\newcommand{\omegavec}{\boldsymbol{\omega}}
\newcommand{\pivec}{\boldsymbol{\pi}}

\newcommand{\defeq}{\coloneqq}

\newcommand{\xvec}{\boldsymbol{x}}
\newcommand{\rhovec}{\boldsymbol{\rho}}
\newcommand{\phivec}{\boldsymbol{\phi}}
\newcommand{\muvec}{\boldsymbol{\mu}}
\newcommand{\pvec}{\boldsymbol{p}}
\newcommand{\yvec}{\boldsymbol{y}}

\newcommand{\Fmat}{\boldsymbol{F}}

\newcommand{\fvec}{\boldsymbol{f}}

\newcommand{\X}{\mathcal{X}}

\newcommand{\cvec}{\boldsymbol{c}}

\newcommand{\inst}[1]{\mathfrak{#1}}
\newcommand{\argmax}{\operatornamewithlimits{argmax}}

	\newcommand{\pers}{\Phi^\diamond}
	
\newcommand{\us}{f} 
\newcommand{\ur}{u} 
\newcommand{\U}{F} 
\newcommand{\V}{U} 
\newcommand{\Rs}{R} 
\newcommand{\Rr}{V} 

\newtheorem{definition}{Definition}

\usepackage{pifont}
\usepackage[normalem]{ulem} 

\definecolor{mygreen}{rgb}{0.0, 0.5, 0.0}
\definecolor{myorange}{rgb}{0.55, 0.62, 1}

\allowdisplaybreaks


%% file: src/abstract.tex
\begin{abstract}
	We study a repeated \emph{information design} problem faced by an informed \emph{sender} who tries to influence the behavior of a self-interested \emph{receiver}.
	We consider settings where the receiver faces a \emph{sequential decision making} (SDM) problem.
	At each round, the sender observes the realizations of random events in the SDM problem.
	This begets the challenge of how to incrementally disclose such information to the receiver to \emph{persuade} them to follow (desirable) action recommendations.
	We study the case in which the sender does \emph{not} know random events probabilities, and, thus, they have to gradually learn them while persuading the receiver.
	We start by providing a non-trivial polytopal approximation of the set of sender's persuasive information structures.
	This is crucial to design efficient learning algorithms. 
	Next, we prove a negative result: no learning algorithm can be persuasive.
	Thus, we relax persuasiveness requirements by focusing on algorithms that guarantee that the receiver's regret in following recommendations grows sub-linearly.
	In the \emph{full-feedback} setting---where the sender observes \emph{all} random events realizations---, 
	we provide an algorithm with $\tilde{O}(\sqrt{T})$ regret for both the sender and the receiver.
	Instead, in the \emph{bandit-feedback} setting---where the sender only observes the realizations of random events actually occurring in the SDM problem---, 
	we design an algorithm that, given an $\alpha \in [1/2, 1]$ as input, ensures $\tilde{O}({T^\alpha})$ and $\tilde{O}( T^{\max \{ \alpha, 1-\frac{\alpha}{2} \} })$ regrets, for the sender and the receiver respectively.
	This result is complemented by a lower bound showing that such a regrets trade-off is essentially tight.
\end{abstract}

%% file: src/introduction.tex
\section{Introduction}
Bayesian persuasion~\citep{kamenica2011bayesian} (a.k.a.~\emph{information design}) is the problem faced by an informed {\em sender} who wants to influence the behavior of a self-interested {\em receiver} via the provision of payoff-relevant information.
This captures the problem of ``who gets to know what'', which is fundamental in all economic interactions.
Thus, Bayesian persuasion is ubiquitous in real-world problems, such as, \emph{e.g.}, online advertising~\cite{bro2012send},
voting~\citep{alonso2016persuading,semipublic,castiglioni2019persuading}, traffic routing~\citep{bhaskar2016hardness,castiglioni2020signaling}, 
security~\citep{rabinovich2015information,xu2016signaling}, auctions~\citep{emek2014signaling, badanidiyuru2018targeting,BacchiocchiCMR022, CastiglioniRM022}, and marketing~\citep{babichenko2017algorithmic,candogan2019persuasion}.

We study Bayesian persuasion in settings where the receiver plays in a \emph{sequential decision making} (SDM) problem.
An SDM problem is characterized by a tree structure made by: \emph{decision} nodes, where the receiver takes actions, and \emph{chance} nodes, in which \emph{partially observable} random events occur.  
The sender perfectly observes the realizations of random events, and their goal is to incrementally disclose the acquired information to induce the receiver towards desirable outcomes.
In order to do so, the sender commits to a \emph{signaling scheme} specifying a probability distribution over action recommendations for the receiver at each decision node.
Specifically, the sender commits to a \emph{persuasive} signaling scheme, meaning that the receiver is incentivized to follow recommendations.
We consider the case of a \emph{farsighted} receiver, meaning that they take into account all the possible future events when deciding whether to deviate or \emph{not} from recommendations at each decision node.

With some notable exceptions (see, \emph{e.g.},~\citep{zu2021learning}), Bayesian persuasion models in the literature make the stringent assumption that both the sender and the receiver know the \emph{prior}, which, in our setting, is defined by the probabilities of random events in the SDM problem.
We relax such an assumption by considering an online learning framework in which the sender, without any knowledge of the prior, repeatedly interacts with the receiver to gradually learn the prior while still being persuasive.

\paragraph{Original contributions.}
Our goal is to design online learning algorithms that are no-regret for the sender, while being persuasive for the receiver.
We start by providing a non-trivial polytopal approximation of the set of sender's persuasive signaling schemes.
This will be crucial in designing efficient (\emph{i.e.}, polynomial-time) learning algorithms, and it also shows how a sender-optimal signaling scheme can be found in polynomial time in the offline version of our problem, which may be of independent interest.
Next, we prove a negative result: without knowing the prior, no algorithm can be persuasive at each round with high probability.
Thus, we relax persuasiveness requirements by focusing on learning algorithms that guarantee that the receiver's regret in following recommendations grows sub-linearly, while guaranteeing the same for sender's regret.
First, we study the \emph{full-feedback} case, where the sender observes the realizations of \emph{all} the random events that may potentially happen in the SDM problem.
In such a setting, we provide an algorithm with $\tilde{O}(\sqrt{T})$ regret for both the sender and the receiver.
Then, we focus on the \emph{bandit-feedback} setting, where the sender only observes the realizations of random events on the path in the tree traversed during the SDM problem.
In this case, we design an algorithm that achieves $\tilde{O}({T^\alpha})$ sender's regret and $\tilde{O}( T^{\max \{ \alpha, 1-\frac{\alpha}{2} \} })$ receiver's regret, for any $\alpha \in [1/2, 1]$ given as input.
The crucial component of the algorithm is a non-trivial exploration phase that uniformly explores the tree defining the SDM problem to build suitable estimators of the prior.
This is needed since, with bandit feedback, playing a signaling scheme may provide insufficient information about its persuasiveness.
Finally, we provide a lower bound showing that the regrets trade off achieved by our algorithm is tight for $\alpha \in [1/2,2/3]$.

%% file: src/related_works.tex
\paragraph{Related works.}
Some works addressed Bayesian persuasion in \emph{Markov decision processes} (MDPs).
\citet{Gan2021Bayesian}~and~\citet{Wu2022Sequential} show how to efficiently find a sender-optimal policy when the receiver is \emph{myopic} (\emph{i.e.}, it only optimizes one-step rewards) in MDPs with infinite and finite horizon, respectively.
Moreover, the former assume that the environment is known, while the latter do \emph{not}.
These works considerably differ from ours, since we assume a farsighted receiver and also model partial observability of random events.\footnote{\citet{Gan2021Bayesian}~also study a model with farsighted receiver, where they show that the problem of finding a sender-optimal policy is \textsf{NP}-hard. Thus, they do \emph{not} provide any algorithmic result for such a model.}
Another work close to ours is~\citep{zu2021learning}, which studies a (non-sequential) persuasion problem in which the sender and the receiver do \emph{not} know the prior and interact online.
\citet{zu2021learning}~provide a persuasive learning algorithm, while, in our model, we show that the ignorance of the prior precludes the possibility of committing to persuasive signaling schemes, and, thus, we need to resort to new techniques to circumvent the issue.
Another line of research, that uses similar techniques as the one employed in this work, studies learning in sequential decision making problems while satisfying unknown constraints~\citep{LucaCFGMT21, LucaCCM0T22}.
Finally,~\citet{Celli2020Private} study Bayesian persuasion with multiple receivers interacting in an imperfect-information sequential game.
Differently from ours, their model adopts a different notion of persuasiveness, known as \emph{ex ante} persuasiveness, and it assumes that the prior is known.
Other works study learning problems in which the sender does \emph{not} know the receivers' payoffs (but knows the prior); see, \emph{e.g.},~\citep{castiglioni2020Online,castiglioni2021Online, CastiglioniM022}.

%% file: src/preliminaries.tex
\section{Preliminaries}

\subsection{Sequential decision making problems}\label{subsec:sdm}

An instance of an SDM problem is defined by a tree structure, utilities, and random events probabilities.
The tree structure has a set of nodes $\Hi \coloneqq \Z \cup \Hi_d \cup \Hi_c $, where: $\Z$ contains all the \emph{terminal nodes} in which the problem ends (corresponding to the leaves of the tree), $\Hi_d$ is the set of \emph{decision nodes} in which the agent acts, while $\Hi_c$ is the set of \emph{chance nodes} where random events occur.
Given any non-terminal node $h \in \Hi \setminus \Z$, we let $A(h)$ be the set of arcs outgoing from $h$.
If $h \in \Hi_d$, then $A(h)$ is the set of receiver's actions available at $h$, while, if $h\in\Hi_c$, then $A(h)$ encodes the possible outcomes of the random event occurring at $h$. 
Furthermore, the utility function $\ur : \Z \to [0,1]$ defines the agent's payoff $\ur(z)$ when the problem ends in terminal node $z \in \Z$.
Finally, each chance node $h \in \Hi_c$ is characterized by a probability distribution $\mu_h \in \Delta_{A(h)}$ over the possible outcomes of the corresponding random event, with $\mu_h(a)$ denoting the probability of action $a \in A(h)$.\footnote{For a finite set $X$ we denote with $\Delta_X$ the set of probability distributions over $X$.}

In an SDM problem, the agent has \emph{imperfect information}, since they do \emph{not} perfectly observe the outcomes of random events. 
Thus, the set of decision nodes $\Hi_d$ is partitioned into \emph{information sets} (infosets for short), where an infoset $I \subseteq \Hi_d$ is a subset of decision nodes that are indistinguishable for the agent.
We denote the set of infosets as $\I$.
For every infoset $I \in \I$ and pair of nodes $h,h^\prime\in I$, it must be the case that $A(h)=A(h^\prime) =: A(I)$, otherwise the agent could distinguish between the two nodes.
We assume that the agent has \emph{perfect recall}, which means that they never forget information once acquired.
Formally, this is equivalent to assume that, for every infoset $I \in \I$, all the paths from the root of the tree to a node $h \in I$ identify the same ordered sequence of agent's actions.
\subsection{Bayesian persuasion in sequential decision making problems}\label{subsec:bpsdm}

We study \emph{Bayesian persuasion in SDM} (BPSDM) problems.
These extend the classical {Bayesian} persuasion framework~\citep{kamenica2011bayesian} to SDM problems by introducing an exogenous agent that acts as a \emph{sender} by issuing signals to the decision-making agent (the \emph{receiver}).\footnote{Appendix \ref{app:bp} shows that BPSDM reduces to classical Bayesian persuasion when there is no sequentiality.}
By following the Bayesian persuasion terminology, the probability distributions $\mu_h$ for each chance node $h$ are collectively referred to as the \emph{prior}.
Thus, the sender observes the realizations of random events occurring in the SDM problem and can partially disclose information to influence the receiver's behavior.
Moreover, the sender has their own utility function defined over terminal nodes, denoted as $\us: \Z \to [0,1]$, and their goal is to commit to a publicly known \emph{signaling scheme} that maximizes their utility in expectation with respect to the prior, the selected signaling scheme, and the receiver's strategy.

Formally, a signaling scheme for the sender defines a probability distribution $\phi_h \in \Delta_{S(h)}$ at each decision node $h \in \Hi_d$, where $S(h)$ is a finite set of signals available at $h$.
During the SDM problem, when the receiver reaches a node $h \in \Hi_d$ belonging to an infoset $I \in \I$, the sender draws a signal $s \sim \phi_h$ and communicates it to the receiver.
Then, based on the history of signals observed from the beginning of the SDM problem ($s$ included), the receiver computes a \emph{posterior} belief over the nodes belonging to the infoset $I$ and plays so as to maximize their expected utility in the SDM sub-problem that starts from $I$, taking into account the just acquired information.

As customary in these settings, a simple revelation-principle-style argument allows us to focus on signaling schemes that are \emph{direct} and \emph{persuasive}~\citep{arieli2019private,kamenica2011bayesian}.
In particular, a signaling scheme is direct if signals correspond to action recommendations, namely $S(h) = A(h)$ for all $h \in \Hi_d$.
A direct signaling scheme is persuasive if the receiver is incentivized to follow action recommendations issued by the sender.
Moreover, we assume that, if the receiver does \emph{not} follow action recommendations at some decision node, then the sender stops issuing recommendations at nodes later reached during the SDM problem.
This is without loss of generality.\footnote{
For a discussion on a similar problem in the field of correlation in sequential games, we refer to~\citep{morrill2021hindsight,von2008extensive}.}

\subsection{The sequence-form representation}\label{subsec:sequence}

The \emph{sequence form} is a commonly-used, compact way of representing \emph{(mixed) strategies} in SDM problems~\citep{koller1996efficient}.
In this work, the sequence-form representation will be employed for receiver's strategies, and to encode the signaling schemes and priors, as we describe in the following. 
\paragraph{Receiver's strategies.}
Given any $h \in \Hi$, we let $\sigma_r(h)$ be the ordered \emph{sequence} of receiver's actions on the path from the root of the tree to node $h$.
By the perfect recall assumption, given any infoset $I \in \I$, it holds that $\sigma_r(h) =\sigma_r(h') =: \sigma_r(I)$ for every pair of nodes $h, h' \in I$.
Thus, we can identify sequences with infoset-action pairs, with $\sigma = (I,a)$ encoding the sequence of actions obtained by appending action $a \in A(I)$ at the end of $\sigma_r(I)$, for any infoset $I \in \I$.
Moreover, $\varnothing$ denotes the \emph{empty sequence}.
Hence, the receiver's sequences are
\(
\Sigma_r \coloneqq  \left\{ (I,a) \mid  I\in\I, a\in A(I) \right\}\cup\{\varnothing\}.
\)
In the sequence-form representation, mixed strategies are defined by specifying the probability of playing each sequence of actions.
Thus, a receiver's strategy is represented by a vector $\xvec \in [0, 1]^{| \Sigma_r |}$,
where $\xvec[\sigma]$ encodes the realization probability of sequence $\sigma \in \Sigma_r$.
Furthermore, a sequence-form strategy is well-defined if and only if it satisfies the following linear constraints:
\begin{align*}
\textstyle{\xvec[\varnothing]=1 \quad \text{and} \quad \xvec[\sigma_r(I)] = \sum_{a\in A(I)}\xvec[\sigma_r(I)a] \quad \forall I \in \I.}
\end{align*}
We denote by $\X_r$ the polytope of all receiver's sequence-form strategies.
We will also need to work with the sets of receiver's strategies in the SDM sub-problem that starts from an infoset $I \in \I$, formally defined as
\(
	\X_{r,I} \defeq \left\{  \xvec \in \X_r \mid \xvec[\sigma_r(I)] = 1 \right\} .
\)

\paragraph{Signaling schemes.}
We represent signaling schemes in sequence form by leveraging the fact that the sender can be thought of as a perfect-information agent who plays at the decision nodes of the SDM problem, since their actions correspond to recommendations for the receiver.
Thus, since sender's infosets correspond to decision nodes, their sequences 
\(
\Sigma_s \defeq \left\{ (h,a) \mid  h\in\Hi_d, a\in A(h) \right\}\cup\{\varnothing\}.
\)
Then, we denote the polytope of (sequence-form) signaling schemes as $\Phi \subseteq [0,1]^{|\Sigma_s|}$, where each signaling scheme is represented as a vector $\phivec\in[0,1]^{|\Sigma_s|}$ satisfying:
\[
\textstyle{\phivec[\varnothing]=1 \quad \text{and} \quad \phivec[\sigma_s(h)]=\sum_{a\in A(h)}\phivec[\sigma_s(h)a] \quad \forall h\in\Hi_d,}
\]
where, similarly to $\sigma_r(h)$ for the receiver, $\sigma_s(h)$ denotes the sender's sequence identified by $h\in\Hi$.
We also define $\Pi \coloneqq \Phi \cap \{0,1\}^{|\Sigma_s|}$ as the set of \emph{deterministic} signaling schemes, which are those that recommend a single action with probability one at each decision node.

\paragraph{Priors.}
We also encode prior probability distributions $\mu_h$ by means of the sequence form.
Indeed, these can be though of as elements of a fixed strategy played by a (fictitious) perfect-information agent that acts at chance nodes.
Thus, for such a chance agent, we define $\Sigma_c$, $\X_c$, and $\sigma_c(h)$ as their counterparts previously introduced for the receiver.
Moreover, in the following, we denote by $\muvec^\star \in \X_c$ the (sequence-form) prior, recursively defined as follows:
\[
\textstyle{\muvec^\star[\varnothing] \coloneqq 1 \quad \text{and} \quad \muvec^\star[\sigma_c(h) a] \coloneqq \muvec^\star[\sigma_c(h)] \, \mu_h(a) \quad \forall h \in \Hi_c, \forall a \in A(h).}
\]

\paragraph{Ordering of sequences.}
For the sake of presentation, we introduce a partial ordering relation among sequences.
Given two sequences $\sigma = (I,a) \in \Sigma_r$ and $\sigma' = (J,b) \in \Sigma_r$, we write $\sigma \preceq \sigma'$ (read as $\sigma$ \emph{precedes} $\sigma'$), whenever there exists a path in the tree connecting a node in $I$ to a node in $J$, and such a path includes action $a$.
We adopt analogous definitions for sequences in $\Sigma_s$ and $\Sigma_c$.\footnote{We refer the reader to Appendix~\ref{sec_app:simple_example} for an example of SDM problem and its sets of sequences.}

%% file: src/online_problem.tex
\section{Learning to persuade}\label{sec:learningtopersuade}

%
In this work, we relax the strong assumption that both the sender and the receiver know the prior $\muvec^\star$ by casting the BPSDM problem into an \emph{online learning framework} in which the sender repeatedly interacts with the receiver over a time horizon of length $T$.
At each round $t \in [T]$, the interaction goes as follows:\footnote{Throughout this work, for $n\in\mathbb{N}$, we denote with $[n]$ the set $\{1,\ldots, n\}$.}
	(i) the sender commits to a signaling scheme $\phivec_t \in \Phi$;
	%
	%
	(ii) a vector $\yvec_t \in \{0,1\}^{|\Sigma_c|}$ encoding realizations of random events is drawn according to $\muvec^\star$;
	%
	(iii) the sender and the receiver play an instance of the (one-shot) BPSDM problem (detailed in Section~\ref{subsec:bpsdm}), in which the sender commits to $\phivec_t$, random events at chance nodes are realized as defined by $\yvec_t$, and the receiver sticks to the recommendations issued by the sender;
	and (iv) the sender observes a \emph{feedback} on the realization of random events at chance nodes, which can be of two types: \emph{full feedback} when the sender observes $\yvec_t$, which specifies the realizations of \emph{all} the random events at chance nodes that are possibly reachable during the SDM problem; \emph{bandit feedback} when the sender observes the terminal node $z_t \in \Z$ reached at the end of the SDM problem.
	The latter is equivalent to observing the realizations of random events at the chance nodes that are actually reached during the SDM problem, namely $\sigma_c(z_t)$.
%
%
%

By letting $\pers(\muvec^\star)$ be the set of persuasive signaling schemes, \emph{i.e.}, such that the receiver is incentivized to following recommendations (a formal definition is provided in Definition~\ref{def:persuasiveness}),
the goal of the sender is to select a sequence of signaling schemes, namely $\phivec_1, \ldots, \phivec_T$, which maximizes their expected utility, while guaranteeing that each signaling scheme $\phivec_t$ is persuasive, namely $\phivec_t \in \pers(\muvec^\star)$.
%

We measure the performance of a sequence $\phivec_1, \ldots, \phivec_T$ of signaling schemes by comparing it with an optimal (fixed) persuasive signaling scheme.
Formally, given a signaling scheme $\phivec \in \Phi$, we first define $\V(\phivec,\muvec^\star)$, respectively $\U(\phivec,\muvec^\star)$, as the expected utility achieved by the receiver, respectively the sender, whenever the former follows action recommendations.
These can be expressed as linear functions of $\phivec$, which, for any $\muvec \in \X_c$, are defined as follows:
\[
	\V(\phivec,\muvec) \defeq \sum\limits_{z\in\Z} \muvec[\sigma_c(z)]\phivec[\sigma_s(z)]\ur(z), \quad \U(\phivec,\muvec) \defeq \sum\limits_{z\in\Z} \muvec[\sigma_c(z)]\phivec[\sigma_s(z)]\us(z).
\]
Finally, by letting $\phivec^\star \in \argmax_{\phivec\in\pers(\muvec^\star)} \U(\phivec, \muvec^\star)$ be an optimal (fixed) persuasive signaling scheme, the sender' performance over $T$ rounds is measured by the \emph{(cumulative) sender's regret}:
\[
	{
	\Rs_T \defeq \sum_{t\in[T]} \Big( \U(\phivec^\star,\muvec^\star)- \U(\phivec_t,\muvec^\star) \Big).
}
\]
%
%
The goal is to design learning algorithms (for the sender) which select sequences of persuasive signaling schemes such that $\Rs_T$ grows asymptotically sub-linearly in $T$, namely $R_T = o(T)$.

%% file: src/characterization.tex
\section{On the characterization of persuasive signaling schemes}\label{sec:known_prior}
\subsection{A local decomposition of persuasiveness}\label{subsec:decomp_pers}

In this section, we formally introduce the set of persuasive signaling schemes $\pers(\muvec^\star)$ as the set of signaling schemes for which the receiver's expected utility by following recommendations is greater than the one provided by an optimal \emph{deviation policy} (DP).\footnote{For ease of exposition, all the definitions and results in this section are provided for the prior $\muvec^\star$. It is straightforward to generalize them to the case of a generic $\muvec \in \X_c$.}
In addition, we show how to decompose any DP into components defined locally at each infoset, which will be crucial in the following Section~\ref{subsec:pers_polytope}.
Intuitively, a DP for the receiver is specified by two elements: (i) a set of \emph{deviation points} in which the DP prescribes to stop following action recommendations; and (ii) the \emph{continuation strategies} to be adopted after deviating from recommendations.

We represent deviation points by vectors $\omegavec\in\{0,1\}^{|\Sigma_r|} $, which are defined so that $\omegavec[\sigma]=1$ if and only if the DP prescribes to deviate upon observing the sequence of action recommendations $\sigma \in \Sigma_r$.
Moreover, by leveraging the w.l.o.g. assumption that the sender stops issuing recommendations after the receiver deviated from them, we focus on DPs such that each path from the root of the tree to a terminal node involves only one deviation point.
As a result, the set of all valid vectors $\omegavec\in\{0,1\}^{|\Sigma_r|} $ is formally defined as
\(
\Omega \defeq \left\{\omegavec\in\{0,1\}^{|\Sigma_r|}  \, \big| \,  \sum_{\substack{\sigma\in\Sigma_r :  \sigma\preceq\sigma_r(z)}}\omegavec[\sigma]\le1 \quad \forall z\in\Z \right\}.
\)

We represent the continuation strategies of DPs by introducing the set of \emph{continuation strategy profiles}, denoted as $\mathcal{P} \defeq \bigtimes_{\sigma = (I,a)\in\Sigma_r}\X_{r, I}$.
A continuation strategy profile $\rhovec \in \mathcal{P}$, with $\rhovec = (\rhovec_\sigma)_{\sigma\in\Sigma_r}$, defines a strategy $\rhovec_\sigma\in \X_{r,I}$ for every receiver's sequence $\sigma=(I,a) \in \Sigma_r$.
Intuitively, $\rhovec_\sigma$ is the strategy for the SDM sub-problem starting from infoset $I$ that is used by the receiver after deviating upon observing sequence $\sigma \in \Sigma_r$.
As a result, any pair $(\omegavec, \rhovec) \in \Omega \times \mathcal{P}$ specifies a valid DP; formally:
\begin{definition}[Deviation policy]
	Given a vector $\omegavec \in\Omega$ and a profile $\rhovec \in \mathcal{P}$, the \emph{$(\omegavec, \rhovec)$-DP} prescribes to follow sender's recommendations until action $a$ is recommended at infoset $I$ for some sequence $\sigma = (I, a) $ such that $\omegavec[\sigma] = 1$; from that point on, it prescribes to play according to strategy $\rhovec_\sigma$.
\end{definition}

We denote by $\V^{\omegavec\to\rhovec}(\phivec,\muvec^\star)$ the receiver's expected utility obtained with a $(\omegavec, \rhovec)$-DP,
so that we can state the following formal definition of persuasive signaling schemes. 
\begin{definition}[Persuasiveness]\label{def:persuasiveness}
	A signaling scheme $\phivec \in \Phi$ is \emph{$\epsilon$-persuasive}, namely $\phivec \in \pers_\epsilon(\muvec^\star)$, if
	\begin{equation}\label{eq:eps_persuasive}
		\max\limits_{(\omegavec,\rhovec)\in\Omega\times \mathcal{P}}\V^{\omegavec\to\rhovec}(\phivec,\muvec^\star)- \V(\phivec,\muvec^\star)\le \epsilon.
	\end{equation}
	Moreover, a signaling scheme $\phivec \in \Phi$ is \emph{persuasive}, namely $\phivec \in \pers(\muvec^\star)$, if it is $0$-persuasive.
\end{definition}
Intuitively, the above definition states that a signaling scheme is $\epsilon$-persuasive if the receiver's expected utility by following recommendations is at most $\epsilon$ less than the one obtained by an optimal DP, which is a DP maximizing receiver's expected utility.

Our local decomposition of DPs is based on suitably-defined, simple deviation policies, which we call \emph{single-point DPs} (SPDPs).
These are a special case of DPs that stop following sender's action recommendations only when a specific single infoset is reached and a particular action is recommended therein.
\footnote{SPDPs are based on the idea of \emph{trigger agents}, which have been originally introduced for computing correlated equilibria in sequential games~\citep{celli2020no,dudik2012sampling}.}
SPDPs are formally defined as follows:
\begin{definition}[Single-point deviation strategy]
	Given a receiver's sequence $\sigma=(I,a) \in \Sigma_r$ and a receiver's strategy $\rhovec_\sigma \in\X_{r,I}$ for the SDM sub-problem starting from infoset $I$, the \emph{$(\sigma,\rhovec_\sigma)$-SPDP} prescribes to follow sender's recommendations until action $a$ is recommended at infoset $I$; from that point on, the strategy prescribes to play according to $\rhovec_\sigma$. 
\end{definition}
We denote by $\V_{\sigma\to \rhovec_\sigma }(\phivec,\muvec^\star)$ the receiver's expected utility obtained by following an $(\sigma,\rhovec_\sigma)$-SPDP.

The following theorem provides the key result underlying our decomposition.\footnote{All the proofs are provided in the Appendices~\ref{sec_app:proof_4},~\ref{sec_app:proof_5},~\ref{app:full_feed},~and~\ref{sec_app:proof_7}.}
It shows that the difference between the utility achieved by a $(\omegavec,\rhovec)$-DP and that obtained by following recommendations can be decomposed into the sum over all the sequences $\sigma \in \Sigma_r$ of analogous differences defined for the $(\sigma,\rhovec_\sigma)$-SPDPs, where each difference is weighted by $\omegavec[\sigma]$.
\begin{restatable}{thm}{decomposition}
	\label{th:decomposition_zin}
	Given a signaling scheme $\phivec \in \Phi$ and a $(\omegavec, \rhovec)$-DP, it holds:
	\[
		\V^{\omegavec\to\rhovec}(\phivec,\muvec^\star) - \V(\phivec,\muvec^\star) = \sum_{\sigma\in\Sigma_r}\omegavec[\sigma] \Big(\V_{\sigma\to\rhovec_\sigma}(\phivec,\muvec^\star) - \V(\phivec,\muvec^\star)\Big).
	\]
\end{restatable}

%% file: src/offline.tex
\subsection{A polytopal approximation of the set of persuasive signaling schemes}\label{subsec:pers_polytope}
In the following, we show how to exploit Theorem~\ref{th:decomposition_zin} to provide an approximate characterization of the set $\pers_\epsilon(\muvec^\star)$ using a polynomially-sized polytope.
First, we state a corollary of Theorem~\ref{th:decomposition_zin} showing that persuasiveness can be bounded by suitably defined SPDPs.
Formally:\footnote{Given any $x \in \mathbb{R}$, we let $[x]^+ \coloneqq \max(x, 0)$.}
\begin{restatable}[]{cor}{corollaryone}
	\label{cor:decompostion_regret}
	Given a signaling scheme $\phivec\in\Phi$, the following holds:
	\[
	\max\limits_{(\omegavec,\rhovec)\in\Omega\times \mathcal{P}}\V^{\omegavec\to\rhovec}(\phivec,\muvec^\star)- \V(\phivec,\muvec^\star)\le \sum\limits_{\sigma= (I,a)\in\Sigma_r} \left[\max\limits_{\rhovec_\sigma \in\X_{r,I}} \V_{\sigma\to\rhovec_\sigma}(\phivec,\muvec^\star)-\V(\phivec,\muvec^\star)\right]^+.
	\]
\end{restatable} 

By exploiting Corollary~\ref{cor:decompostion_regret}, we introduce the following definition of \emph{$\epsilon$-persuasive polytope} (Lemma~\ref{thm:polytope} justifies the term polytope), as the set of signaling schemes for which there is no $(\sigma,\rhovec_\sigma)$-SPDP that achieves a receiver's utility that exceeds by more than $\epsilon / |\Sigma_r|$ that of following recommendations.

\begin{definition}[Persuasive polytope]\label{def:pers_poly}
	The \emph{$\epsilon$-persuasive polytope} is defined as:
	\[
	\Lambda_\epsilon(\muvec^\star) \coloneqq \Big\{\phivec\in \Phi\, \Big\vert\, \max_{\rhovec_\sigma \in \X_{r,I}} \V_{\sigma\to \rhovec_\sigma}(\phivec,\muvec^\star)-\V(\phivec,\muvec^\star)\le {\epsilon/ |\Sigma_r|}\quad \forall \sigma\in\Sigma_r\Big\}.
	\]
	Moreover, we denote by $\Lambda(\muvec^\star)$ the $0$-persuasive polytope.
\end{definition}

As we show in the following lemma, $\Lambda_\epsilon(\muvec^\star)$ is an efficiently-representable polytope.
\begin{restatable}[]{lem}{polytope}\label{thm:polytope}
	The set $\Lambda_\epsilon(\muvec^\star)$ can be described by means of a polynomial number of linear constraints.
\end{restatable}

The following lemma   shows that the $\epsilon$-persuasive polytope is contained in $\pers_\epsilon(\muvec^\star)$, and that the set of persuasive signaling schemes is contained in the former.
Formally:
\begin{restatable}[]{lem}{containment}\label{thm:containment}
	It is always the case that $\pers(\muvec^\star) \equiv \Lambda(\muvec^\star) \subseteq \Lambda_\epsilon(\muvec^\star) \subseteq \pers_\epsilon(\muvec^\star)$.
\end{restatable}

Lemma~\ref{thm:containment} also implies that the polytope $\Lambda(\muvec^\star)$ exactly characterizes the set of persuasive signaling schemes $\pers(\muvec^\star)$. 
Thus, by adding the maximization of the sender's expected utility $F(\phivec,\muvec^\star)$ on top of the linear constraints describing $\Lambda(\muvec^\star)$, we obtain a polynomially-sized linear program for finding an optimal sender's signaling scheme in any instance of the BPSDM problem in which $\muvec^\star$ is known.
\begin{restatable}[]{thm}{offline}\label{thm:offline}
	The BPSDM problem can be solved in polynomial time when the prior $\muvec^\star$ is known.
\end{restatable}

%% file: src/online_learning.tex
\section{Always being persuasive is impossible: a relaxation is needed}\label{sec:online}

In this section, we prove that it is impossible to design an algorithm that returns a sequence of persuasive signaling schemes for a generic BPSDM problem.
Motivated by this result, we introduce a new online learning problem that relaxes persuasiveness requirements.

First, we provide the following impossibility result:
\begin{restatable}[Impossibility of persuasiveness]{thm}{impossibility}
	\label{th:impossibility}
	There exists a constant $\gamma \in (0,1)$ such that no algorithm can guarantee to output a sequence $\phivec_1, \ldots,\phivec_T$ of signaling schemes such that, with probability al least $\gamma$, all the signaling schemes $\phivec_t$ are persuasive.
\end{restatable}
Notice that this result is in contrast with what happens in the basic case of non-sequential Bayesian persuasion (see the work by~\citet{zu2021learning}), where it is possible to design no-regret algorithms that output sequences of signaling schemes that are guaranteed to be persuasive with high probability.

Theorem~\ref{th:impossibility} motivates the introduction of a less restrictive requirement on the signaling schemes output by a learning algorithm.
In particular, we look for algorithms that output sequences $\phivec_1, \ldots, \phivec_T$ of signaling schemes such that the expected utility loss incurred by the receiver by following sender's recommendations rather than playing an optimal DP is small.
To capture such a requirement, we introduce the following definition of \emph{(cumulative) receiver's regret}:
\[
	\Rr_T \coloneqq \sum\limits_{t\in[T]} \max\limits_{(\omegavec,\rhovec)\in\Omega\times \mathcal{P}}\V^{\omegavec\to\rhovec}(\phivec_t,\muvec^\star) -\sum\limits_{t\in[T]} \V(\phivec_t,\muvec^\star).
\]
Therefore our goal becomes that of designing algorithms guaranteeing that the cumulative receiver's regret grows sub-linearly in $T$, namely $\Rr_T = o(T)$, while continuing to ensure that $\Rs_T = o(T)$.

In Sections~\ref{sec:full_feedback}~and~\ref{sec:unkown_prior}, we design algorithms achieving sub-linear $\Rr_T$ and $\Rs_T$ for the learning problem described in Section~\ref{sec:learningtopersuade}.
The algorithms implement two functions:
	(i) $\textsc{SelectStrategy}()$, which, at each $t \in [T]$, draws a signaling scheme $\phivec_t \in\Phi$ on the basis of the internal state of the algorithm; and
	(ii) $\textsc{Update}(o_t)$, which modifies the internal state on the basis of the observation $o_t$ received as feedback.
Each algorithm alternates these two functions as the interaction between the sender and the receiver unfolds as described in Section~\ref{sec:learningtopersuade}. Specifically, under full feedback the sender observes $\yvec_t$ and calls $\textsc{Update}(\yvec_t)$, while in the bandit feedback it observes $z_t$ and calls $\textsc{Update}(z_t)$.

%% file: src/full_feedback.tex
\section{Learning with full feedback}\label{sec:full_feedback}
\begin{wrapfigure}[13]{R}{0.32\textwidth}
	\begin{minipage}{0.32\textwidth}
		\vspace{-0.6cm}
		\begin{algorithm}[H]
			\small
			\caption{Full-feedback algorithm}
			\begin{algorithmic}
				\Function{\textsc{SelectStrategy}}{\hspace{0cm}}:
				\State $\phivec_t\gets \arg\max\limits_{\phivec\in\Lambda_{\beta_t}(\widehat\muvec_t)} \U(\phivec, \widehat \muvec_t)$
				\State\Return $\phivec_t$
				\EndFunction
				\\\hrulefill
				\Function{\textsc{Update}}{$\yvec_t$}:
				\State $\widehat\muvec_{t+1}[\sigma]\gets \frac{1}{t}\sum\limits_{\tau=1}^t \yvec_\tau[\sigma]\,\,\forall\sigma\in \Sigma_c$
				\State $\epsilon_{t+1}\gets\sqrt{\frac{\log(2T|\Sigma_c|/\delta)}{2t}}$
				\State $\beta_{t+1}\gets 2|\Z||\Sigma_r|\epsilon_{t+1}$
				\EndFunction
			\end{algorithmic}%
			\label{alg:full_feedback}
		\end{algorithm}
	\end{minipage}
\end{wrapfigure}
We start by providing a learning algorithm (Algorithm~\ref{alg:full_feedback}) working with full feedback, \emph{i.e.}, when the sender observes the realizations of \emph{all} the possible random events.
The main idea of the algorithm is to choose signaling schemes $\phivec_t$ that belong to suitable sets $\Lambda_{\beta_t}(\widehat\muvec_t)$ which are designed to be ``close'' to the set $\pers(\muvec^\star)$ of persuasive signaling schemes.
At each round $t \in [T]$, Algorithm~\ref{alg:full_feedback} defines the desired set as follows.
First, it maintains an estimate $\widehat\muvec_t$ of $\muvec^\star$; formally, it defines a radius $\epsilon_t$ such that the event $\mathcal{E}\coloneqq\left\{\|\widehat\muvec_t-\muvec^\star\|_\infty\le\epsilon_t \,\,\,\forall t\in[T]\right\}$ holds with probability at least $1-\delta$.
Second, it defines a parameter $\beta_t$ such that, conditionally to the realization of the event $\mathcal{E}$, the following two conditions hold: (i) the decision space $\Lambda_{\beta_t}(\widehat\muvec_t)$ contains the optimal signaling scheme $\phivec^\star$; (ii) $\Lambda_{2\beta_t}(\muvec^\star)$ contains the signaling scheme $\phivec_t$.
Intuitively, the first condition is needed to have low sender's regret, while the second one yields signaling schemes that are approximately persuasive.\footnote{See Lemma~\ref{lem:lemma1}~and~\ref{lem:lemmaunofull} in Appendix~\ref{app:full_feed} for the formal statements of these properties.
}

The polytopal approximation that we provide in Section~\ref{subsec:pers_polytope} plays a crucial role in the complexity of Algorithm~\ref{alg:full_feedback}.
Specifically, it allows it to select the desired $\phivec_t$ in polynomial time by optimizing over the set $\Lambda_{\beta_t}(\widehat\muvec_t)$, which can be done efficiently.
The use of the set $\Lambda_{\beta_t}(\widehat\muvec_t)$ over $\pers_{\beta_t}(\widehat\muvec_t)$ is necessary due to the fact that the latter is  \emph{not} known to admit an efficient representation. Formally:
\begin{restatable}[]{thm}{onlineAlg}
	\label{th:regret_full}
	Given any $\delta\in(0,1)$, with probability at least $1-\delta$, Algorithm~\ref{alg:full_feedback} guarantees:
	\[
	\Rs_T=\mathcal{O}\left( |\Z|\sqrt{T\log(T|\Sigma_c|/\delta)}\right), \quad
	\Rr_T=\mathcal{O}\left(|\Sigma_r||\Z|\sqrt{T\log(T|\Sigma_c|/\delta)}\right).
	\]
	Moreover, the algorithm runs in polynomial time.
\end{restatable}

%% file: src/bandit_feedback.tex
\section{Learning with bandit feedback}\label{sec:unkown_prior}

\begin{wrapfigure}[]{R}{0.45\textwidth}
	\begin{minipage}{0.44\textwidth}
		\vspace{-1.3cm}
		\begin{algorithm}[H]
			\small
			\caption{Bandit-feedback algorithm}
			\begin{algorithmic}
				\Function{\textsc{SelectStrategy}}{\hspace{0cm}}:
				\If{$t\le N$} \textcolor{gray}{ \Comment{First Phase}}
				\State $\sigma=(h,a)\gets \arg\min_{\sigma\in\Sigma_c}C_t[\sigma]$
				\State $\Sigma_s \ni \sigma^\prime \gets \sigma_s(h)$
				\State Choose $\phivec_t \in \Phi: \phivec_t[\sigma^\prime]=1$
				\Else \textcolor{gray}{ \Comment{Second Phase}}
				\State $\phivec_t\gets \arg\max\limits_{\phivec\in\Lambda_{\beta_N}(\widehat\muvec_N)}\max\limits_{\muvec\in \C_t(\delta)} \U(\phivec, \muvec)$
				\EndIf
				\State\Return $\phivec_t$
				\EndFunction
				\\\hrule
				\Function{\textsc{Update}}{$z_t$}:
				\State Build path $\pvec_t\in\{0,1\}^{|\Sigma_c|}$ from $\sigma_c(z_t)$
				\State Sample $\pivec_t\sim\phivec_t$ s.t. $\pvec_t[\sigma]=1 \Rightarrow \sigma \in \Sigma_\downarrow(\pivec_t)$
				\For{$\sigma\in\Sigma_\downarrow(\pivec_t)$}
				\State $C_{t+1}[\sigma]\gets C_{t}[\sigma]+1$
				\State $\widehat\muvec_{t+1}[\sigma]\gets\frac{1}{C_{t+1}[\sigma]}\sum_{\tau=1}^{C_{t+1}[\sigma]}\pvec_\tau[\sigma]$
				\State $\epsilon_{t+1}[\sigma]\gets \sqrt{\frac{\log(4T|\Sigma_c|/\delta)}{2C_{t+1}[\sigma]}}$
				\EndFor
				\State $\C_{t+1}(\delta)\gets$ ${\scriptsize \left\{\muvec \,\Big\vert\, |\muvec[\sigma]-\widehat\muvec_{t+1}[\sigma]|\le\epsilon_{t+1}[\sigma]\, \forall\sigma\in\Sigma_c\right\}}$
				\State $\beta_{t+1}\gets 2|\Z| |\Sigma_c|\sqrt{\frac{|\Sigma_c|\log(4T|\Sigma_c|/\delta)}{2(t+1)}}$
				\EndFunction
			\end{algorithmic}%
			\label{alg:bandit_feedback}
		\end{algorithm}
	\end{minipage}
\end{wrapfigure}
In this section, we build on Algorithm~\ref{alg:full_feedback} to deal with bandit feedback, \emph{i.e.}, when at each round $t \in [T]$ the sender only observes the terminal node $z_t$ reached at the end of the SDM problem.
The main difficulties of such a setting can be summarized by the following observations.
First, the feedback $z_t$ only reveals partial information about the prior, and such information also depends on the selected signaling scheme $\phivec_t$.
Second, even if the sender plays a signaling scheme $\phivec \in \Phi$ for an arbitrarily large number of rounds, there is no guarantee that they collect enough information to tell whether $\phivec \in \pers_\epsilon(\muvec^\star)$ or \emph{not} for some $\epsilon > 0$.
Indeed, the persuasiveness of a signaling scheme depends on \emph{all} receiver's utilities in the SDM problem, and some parts of the tree may \emph{not} be reached during a sufficiently large number of rounds by committing to $\phivec$.
Thus, any algorithm for the bandit-feedback setting must guarantee a suitable level of exploration over the entire tree, so as to keep track of the entity of the violation of persuasiveness constraints.

We design a two-phase algorithm, whose pseudo-code is provided in Algorithm~\ref{alg:bandit_feedback}.
The algorithm takes as input the number $N \in [T]$ of rounds devoted to the \emph{first phase} guaranteeing the necessary amount of {exploration}, as detailed in Section~\ref{subsec:estimating}.
During this phase, the $\textsc{SelectStrategy}()$ procedure implements an efficient deterministic uniform exploration policy, which builds an unbiased estimator $\widehat \muvec_N$ of $\muvec^\star$.
This allows to restrict the space of feasible signaling schemes used in the subsequent phase to those that are approximately persuasive, \emph{i.e.}, those in the set $\Lambda_{\beta_N}(\widehat \muvec_N)$. 
In Section~\ref{subsec:min_regret}, we discuss the \emph{second phase} of the the algorithm, composed by the rounds $t > N$, during which the algorithm focuses on the minimization of sender's regret by exploiting the \emph{optimism in face of uncertainty} principle.
Finally, in Section~\ref{sec:lower_bound}, we provide a lower bound on the trade-off between sender's and receiver's regrets, matching the upper bounds achieved by Algorithm~\ref{alg:bandit_feedback} for a large portion of the trade-off frontier. This result formally motivates the necessity of the uniform exploration which is performed in the first phase of the algorithm.

\subsection{Minimizing the receiver's regret}\label{subsec:estimating}

At each round $t \in [T]$, the sender observes a terminal node $z_t \in \Z$ that uniquely determines a path in the tree defining the SDM problem.
We encode such a path by means of a vector $\pvec_t\in\{0,1\}^{|\Sigma_c|}$ such that $\pvec_t[\sigma] = 1$ if and only if the chance sequence $\sigma\in\Sigma_c$ lies on the path from the root of the tree to $z_t$, namely $\sigma \preceq \sigma_c(z_t)$.
If the sender commits to a signaling scheme $\phivec_t \in \Phi$, then it is easy to see that, for every $\sigma = (h,a) \in \Sigma_c$, the element $\pvec_t[\sigma]$ is distributed as a Bernoulli of parameter $\phivec_t[\sigma_s(h)]\muvec^\star[\sigma]$.
The crucial observation behind the design of our estimator is that, if the sender commits to a deterministic signaling schemes $\pivec_t \in \Pi$ at some round $t \in [T]$, then for all the chance sequences $\sigma\in\Sigma_c$ that are \emph{compatible} with $\pivec_t$, \emph{i.e.}, that can be observed when $\pivec_t$ is played, we have that $\pvec_t[\sigma]$ is distributed as a Bernoulli of parameter $\muvec^\star[\sigma]$.
Formally,  a sequence $\sigma\in\Sigma_c$ is compatible with  $\pivec_t$ if there exists a chance node $h \in \Hi_c$ and an outcome $a \in A(h)$ satisfying $\sigma = (h,a)$ and $\pivec_t[\sigma_s(h)]=1$.
This observation leads to the following result:
\begin{restatable}{lem}{lemmaunbiased}\label{lem:unbiased}
	For every deterministic signaling scheme $\pivec \in \Pi$, let
	\[
	\Sigma_\downarrow(\pivec)\coloneqq \left\{\sigma=(h,a)\in\Sigma_c \mid a\in A(h) \wedge \pivec[\sigma_s(h)]=1 \right\}.
	\]
	Then, during each round $t \leq N$ of Algorithm~\ref{alg:bandit_feedback}, it holds $\mathbb{E}\left[ \pvec_t[\sigma] \right] = \muvec^\star[\sigma]$ for every $\sigma\in\Sigma_\downarrow(\pivec_t)$.
\end{restatable}
Thus, during the first phase, Algorithm~\ref{alg:bandit_feedback} builds the desired estimator $\widehat{\muvec}_N$ of $\muvec^\star$ as follows.
At each round $t \leq N$, after observing the feedback $z_t$, the algorithm samples a deterministic signaling scheme $\pivec_t \in \Pi$ according to $\phivec_t$ (the one actually selected at $t$), so that all the sequences $\sigma \in \Sigma_c$ such that $\pvec_t[\sigma]=1$ (or, equivalently, $\sigma \preceq \sigma_c(z_t)$) belong to $\Sigma_\downarrow(\pivec_t)$.\footnote{The sampling can be done efficiently by a straightforward modification of the recursive procedure in~\cite{farina2021simple,farina2021bandit}.}
Then, for every $\sigma\in\Sigma_\downarrow(\pivec_t)$, the algorithm updates the estimator component $\widehat\muvec_t[\sigma]$ according to $\pvec_t[\sigma]$.
Since the probability of visiting a sequence $\sigma\in\Sigma_c$ depends on $\phivec_t$ (and, thus, can be arbitrarily small), the first $N$ rounds must be carefully used to ensure that each sequence is explored at least $N/|\Sigma_c|$ times.
To explore a specific sequence $\sigma\in\Sigma_c$, we choose a signaling scheme $\phivec_t$ such that $\sigma \in \Sigma_\downarrow(\pivec_t)$ for every deterministic $\pivec_t\sim\phivec_t$.
The procedure described above is needed for minimizing the receiver's regret, since, in the second phase, the algorithm selects signaling schemes $\phivec_t$ from $\Lambda_{\beta_N}(\widehat \muvec_N)$.
In particular, as shown by the following lemma, Algorithm~\ref{alg:bandit_feedback} guarantees that the receiver's regret is upper bounded by $2\beta_N$ at each round $t > N$, since it defines $\epsilon_t[\sigma]$ for each sequence $\sigma \in \Sigma_c$ so that the event
\(
{
	\tilde{\mathcal{E}}\coloneqq\{|\muvec^\star[\sigma]-\widehat\muvec_t[\sigma]|\le\epsilon_t[\sigma] \,\, \forall (t,\sigma)\in[T]\times\Sigma_c\}
}
\)
holds with probability at least $1-\delta/2$.
\begin{restatable}{lem}{lemmaunobandit}\label{lem:lemmaunobandit}
	Under the event $\tilde{\mathcal{E}}$, Algorithm~\ref{alg:bandit_feedback} guarantees that $\phivec_t \in \Lambda_{2\beta_N} (\muvec^\star)$ at each round $t > N$.
\end{restatable}

\subsection{Minimizing the sender's regret}\label{subsec:min_regret}

Algorithm~\ref{alg:bandit_feedback} also needs to guarantee small sender's regret.
To do so, we would like that $\phivec^\star$ is a valid pick for the algorithm, \emph{i.e.}, it belongs to $\Lambda_{\beta_N}(\widehat\muvec_t)$.
However, differently from the full-feedback setting,  stopping exploration after the first $N$ round does \emph{not} guarantee optimal rates.
In order to fix this issue, in the second phase, the algorithm selects $\phivec_t$ optimistically by maximizing the sender's expected utility $\U(\phivec, \muvec)$ over both $\phivec \in\Lambda_{\beta_N}(\widehat\muvec_N)$ and $\muvec \in \C_{t}(\delta)$, where $\C_{t}(\delta)$ is a suitably-defined confidence set centered around $\widehat \muvec_t$ such that $\{\muvec^\star\in\C_t(\delta)\}\equiv\tilde{\mathcal{E}}$, and, thus, it holds with high probability.
This guarantees that $\max_{\muvec \in \C_{t}(\delta)}  \U(\phivec^\star, \muvec)\ge \U(\phivec^\star, \muvec^\star)$.
Formally:
\begin{restatable}{lem}{lemmaduebandit}\label{lem:lemmaduebandit}
	If the event $\tilde{\mathcal{E}}$ holds, then, for every round $t>N$, it holds that
	\(
	\phivec^\star\in \Lambda_{\beta_N}(\widehat\muvec_t)
	\) and 
	\(	
	\max_{\muvec \in \C_t(\delta) } \U(\phivec^\star, \muvec) \ge \U(\phivec^\star,\muvec^\star).
	\)
\end{restatable}
Thus, $\U(\phivec_t,  \muvec^\star) \approx \U(\phivec_t,  \widehat \muvec_t) \ge \max_{\muvec \in \C_t(\delta) } \U(\phivec^\star,  \widehat \muvec)  \ge F(\phivec^\star, \muvec^\star)$ holds in the limit, implying that $\U(\phivec_t,  \muvec^\star) $ converges to $\U(\phivec^\star,  \muvec^\star)$ after sufficiently many rounds.
Formally:
\begin{restatable}[]{thm}{onlineAlgBandit}
	\label{th:bandit_regret}
	Given any $\delta\in(0,1)$ and $N\in [T]$, Algorithm~\ref{alg:bandit_feedback} guarantees:
	\[
	\Rs_T=\mathcal{O}\left( N+\sqrt{\log(T|\Sigma_c|/\delta)|\Sigma_c|T}\right)\quad\text{and}\quad
	\Rr_T=\mathcal{O}\left( N+T|\Z|\sqrt{{|\Sigma_c|\log(T|\Sigma_c|/\delta)/N}}\right),
	\]
	with probability at least $1-\delta$.
	Moreover, the algorithm runs in polynomial time.
\end{restatable}
In contrast to the case with full feedback, the optimization problem solved by Algorithm~\ref{alg:bandit_feedback} belongs to the class of bilinear problems, which are \textsf{NP}-hard in general~\cite{hillar2013most}. However, in Theorem~\ref{th:bandit_regret} we prove that our specific problem can be solved in polynomial time.
Furthermore, notice that Theorem~\ref{th:bandit_regret} takes as input the number $N$ of rounds devoted to the first phase.
Given an $\alpha \geq 1/2 $, by choosing any $N = \lfloor T^\alpha \rfloor$ we get bounds of
$\Rs_T = \tilde{\mathcal{O}}(T^\alpha)$ and 
$\Rr_T = \tilde{\mathcal{O}}(T^{\max\{\alpha,1-\frac{\alpha}{2}\}})$.

%% file: src/bandit_due.tex
\subsection{The lower bound frontier}\label{sec:lower_bound}
\begin{wrapfigure}[11]{l}{0.34\textwidth}
	\vspace{-\baselineskip}
	\centering
	\includegraphics[width=0.27\textwidth]{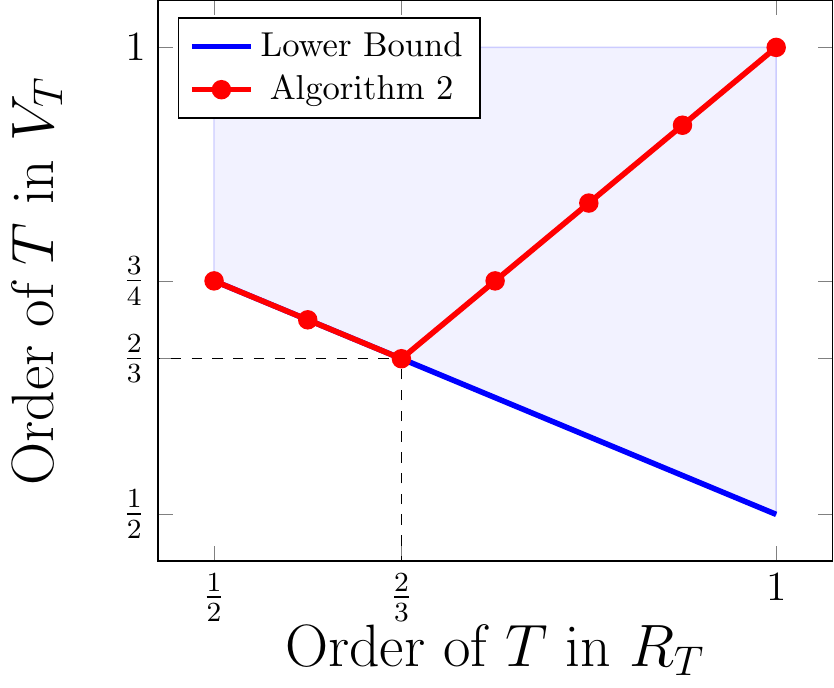}\vspace{-0.0cm}
	\caption{Trade-off between $\Rs_T$ and $\Rr_T$ in the bandit feedback. 
}
\label{fig:tradeoff}
\end{wrapfigure}
We conclude by showing that the trade offs between $\Rr_T$ and $\Rs_T$ achieved by Algorithm~\ref{alg:bandit_feedback} are essentially tight.
Previously, we provided an intuition as to why the algorithm needs to uniformly explore the entire tree of the SDM problem.
Here, we provide a lower bound that corroborates such a statement. 
In particular, the following theorem shows that, for any $\alpha \in  [1/2, \ 1]$, in order to guarantee a sender's regret of the order of $\mathcal{O}(T^{\alpha})$, it is necessary to suffer a receiver's regret of the order of $
{\Omega}(T^{1-\alpha/2})$.\footnote{
For $\alpha \le 1/2$, a simple reduction from a standard multi-armed bandit problem provides a lower bound of $\Omega(\sqrt{T})$ on both sender's regret $\Rs_T$ and receiver's regret $\Rr_T$.}
\begin{restatable}[]{thm}{lowerbound}
\label{th:lower_bound}
For any $\alpha \in [1/2, 1]$, there exists a constant $\gamma \in (0,1)$ such that no algorithm guarantees both $\Rs_T = o(T^\alpha)$ and $\Rr_T = o (T^{1-\alpha/2})$ with probability greater than $\gamma$.
\end{restatable}
Figure~\ref{fig:tradeoff} shows on the horizontal axis the order of the $T$ term in $\Rs_T$, while, on the vertical axis, it shows the order of the $T$ in $\Rr_T$.
The shaded area over the blue line shows the achievable trade offs, while the marked red line shows the performances proved in Theorem~\ref{th:bandit_regret}.
Thus, we show that Algorithm~\ref{alg:bandit_feedback} matches the lower bound for $\alpha \in [1/2,2/3]$.
However, when $\alpha \in [2/3, 1]$, the guarantees proved in Theorem~\ref{th:bandit_regret} diverge from the ones proved in the lower bound.
This is due to the $N = \lfloor T^{\alpha} \rfloor$ component in the receiver's regret that becomes dominant when $\alpha \ge 2/3$.
We conjecture that it is possible to reduce this term to $\sqrt{N}$, hence matching the lower bound of Theorem~\ref{th:lower_bound}.
{The reason for such a gap between the lower and upper bounds is that, during the first phase, Algorithm~\ref{alg:bandit_feedback} utilizes signaling schemes without taking into account their persuasiveness, thus incurring in large receiver's regret during the first steps. We leave as future work addressing the question of whether it is possible to design exploration strategies by only using approximately-persuasive signaling schemes.}

%% file: src/appendix.tex
\input{src/appendix/appendix_example}
\input{src/appendix/appendix_simple_example}
\input{src/appendix/appendix_notation}

\input{src/appendix/appendix_offline}

\input{src/appendix/appendix_online}

\input{src/appendix/appendix_full_feedback}

\input{src/appendix/appendix_bandit_feedback}

%% file: src/appendix/appendix_example.tex
\section{Relation with the classical Bayesian persuasion framework}\label{app:bp}

\begin{wrapfigure}[]{R}{0.54\textwidth}
	\centering
	\includegraphics[width=0.45\textwidth]{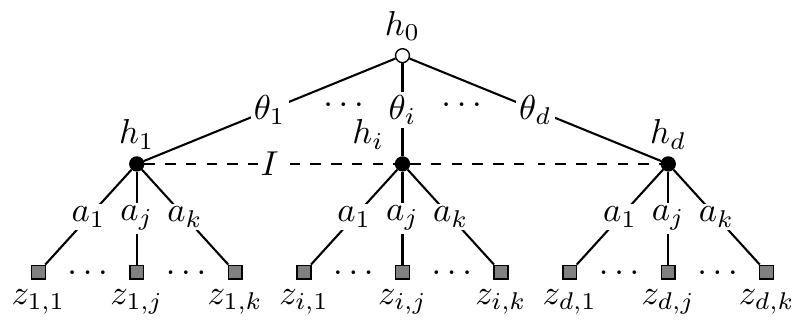}
	\caption{Instance of BPSDM problem corresponding to a given instance of Bayesian persuasion problem.}
	\label{fig:tree_BP}
\end{wrapfigure}

In this section, we clarify the relationship between the BPSDM problems that we study in this paper and the classical Bayesian persuasion framework introduced by~\citet{kamenica2011bayesian}.
In particular, we show that any instance of the classical Bayesian persuasion problem can be mapped to an instance of the BPSDM problem.

A Bayesian persuasion problem instance is defined by a set $\mathcal{A}$ of $k\coloneqq |\mathcal{A}|$ actions for the receiver, a set $\mathcal{S}$ of signals for the sender, and a set $\Theta$ of $d \coloneqq |\Theta|$ possible outcomes of a (single) random event (called \emph{states of nature} in the classical Bayesian persuasion terminology).
The receiver's payoff function is $u^{\textsf{R}}:\Theta\times\mathcal{A}\to[0,1]$, while the sender's one is $u^{\textsf{S}}:\Theta\times\mathcal{A}\to[0,1]$.
The sender observes the realized state of nature, which is drawn according to a commonly-known prior distribution $\mu\in\Delta_\Theta$.
%
%
Then, they partially disclose information about the state by committing to a signaling scheme $\varphi:\Theta\to\Delta_{\mathcal{S}}$, which is a randomized mapping from states of nature to signals for the receiver.
Thus, the interaction between the sender and the receiver is as follows:
\begin{enumerate}[label=(\roman*)]
	\item The sender commits to a publicly known signaling scheme $\varphi$.
	\item The sender observes the state of nature $\theta\sim \mu$.
	\item The sender samples a signal $s\sim \varphi(\theta, \cdot)$ and sends it to the receiver.
	\item The receiver computes their posterior belief over the states $\Theta$.
	\item The receiver plays an action $a\in\mathcal{A}$ that maximizes their expected payoff.
\end{enumerate}
The posterior beliefs that the receiver computes in step (iv) after observing a signal $s \in \mathcal{S}$ are defined by a probability distribution $\xi_s \in \Delta_\Theta$ such that:
\[
\xi_s(\theta) \coloneqq \frac{\mu(\theta)\varphi(\theta, s)}{\sum_{\theta^\prime\in\Theta} \mu(\theta^\prime)\varphi(\theta^\prime, s)},
\]
and, thus, after observing signal $s$ the receiver plays an action
\[
a\in\arg\max\limits_{a^\prime\in\mathcal{A}}\sum\limits_{\theta\in\Theta}\xi_s(\theta) u^{\mathsf{R}}(\theta, a^\prime).
\]

A revelation-principle-style argument~\citep{kamenica2011bayesian} allow the sender to focus on direct and persuasive signaling schemes, where the latter property means that $\mathcal{S}\equiv\mathcal{A}$, with signals corresponding to actions recommendations for the receiver.
%
%
A persuasive signaling scheme $\varphi: \Theta\to\Delta_{\mathcal{S}}$ is such that the receiver is always incentivized to follow action recommendations; formally:
\begin{equation}\label{eq:ic_BP}
	\sum\limits_{\theta\in\Theta}\mu(\theta)\varphi(\theta, a)u^\textsc{R}(\theta, a)\ge\sum\limits_{\theta\in\Theta}\mu(\theta)\varphi(\theta, a)u^\textsc{R}(\theta, a^\prime)\quad\forall a,a^\prime\in\A.
\end{equation}

\paragraph{Instance mapping.}
Given an instance of the classical Bayesian persuasion problem~\citep{kamenica2011bayesian}, a corresponding (equivalent) instance of our BPSDM problem can be constructed as follows:
\begin{enumerate}[label=(\arabic*)]
	\item There is a unique chance node $h_0$ which is the root of the tree defining the SDM problem.
	\item At the chance node, there are $d$ possible outcomes (namely $A(h_0) \equiv \Theta$), each corresponding to a state of nature $\theta\in \Theta$ and having probability $\mu(\theta)$ of occurring, so that with a slight abuse of notation we can write $\muvec^\star[\varnothing] = 1$ and $\muvec^\star[\theta] = \mu(\theta)$ for all $\theta \in \Theta$.
	\item The receiver has a unique infoset $I$, which contains one decision node for each possible outcome at the chance node.
	\item At infoset $I$, the receiver has a set ${A}(I)\equiv\mathcal{A}$ of available actions.
	\item Terminal nodes $\Z$ are determined by  state of nature, receiver's action pairs, so that each $\theta_i\in\Theta$ and $a_j\in\mathcal{A}$ define a corresponding terminal node $z_{i,j}$ in the SDM problem.
\end{enumerate}

The following theorem formally states that our definition of persuasiveness (Definition~\ref{def:persuasiveness}) instantiated to the BPSDM problem instances described above is equivalent to the definition of persuasiveness for classical Bayesian persuasion problems (Equation~\eqref{eq:ic_BP}).
This establishes that our framework encompasses classical Bayesian persuasion problems as a special case.
\begin{restatable}{thm}{BPequalBPSDM}
	Given any Bayesian persuasion instance, a signaling scheme is persuasive (Equation~\eqref{eq:ic_BP}) if and only if it is persuasive (Definition~\ref{def:persuasiveness}) in the corresponding instance of BPSDM problem. 
	%
\end{restatable}
\begin{proof}
	It is sufficient to prove the equivalence between Equation~\eqref{eq:eps_persuasive} for $\epsilon=0$ and Equation~\eqref{eq:ic_BP} applied to the BPSDM problem instance representing the given Bayesian persuasion instance.
	To do that, we employ Theorem~\ref{th:decomposition_zin} and Lemma~\ref{lem:decomposition_1} in such a BPSDM problem instance, so that, using the notation introduced in this section, it is straightforward to see that Equation~\eqref{eq:eps_persuasive} reads as follows:
	\[
	\max\limits_{a^\prime\in \A}\sum\limits_{\theta\in\Theta}\varphi(\theta, a)\mu(\theta)u^{\textsf{R}}(\theta, a^\prime)-\sum\limits_{\theta\in \Theta}\varphi(\theta, a)\mu(\theta)u^{\textsf{R}}(\theta, a)\le 0 \quad \forall a \in \mathcal{A}.
	\]
	%
	By rearranging the terms, we get Equation~\eqref{eq:ic_BP}, which concludes the proof.
\end{proof}

%% file: src/appendix/appendix_simple_example.tex
\section{Example of SDM problem and its sets of sequences}\label{sec_app:simple_example}

\begin{figure}[!htp]
	\centering
	\includegraphics[width=0.5\textwidth]{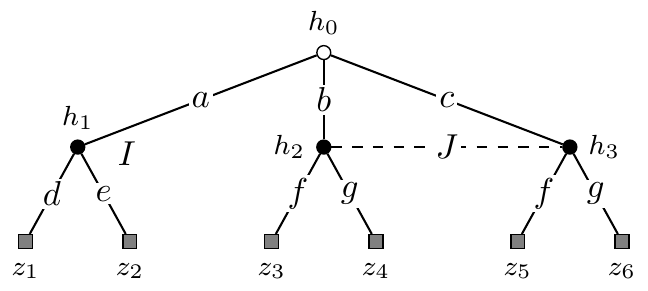}
	\caption{Example of SDM problem and its sets of sequences $\Sigma_r$, $\Sigma_s$, and $\Sigma_c$.}
	\label{fig:tree_example}
\end{figure}

Figure~\ref{fig:tree_example} shows a simple instance of a SDM problem.
This is defined by a tree whose set of chance nodes is $\Hi_c=\{h_0\}$, while the set of decision nodes is $\Hi_d=\{h_1,h_2,h_3\}$.
The set of terminal nodes is $\Z=\{z_1,\ldots,z_6\}$.
Moreover, the set of decision nodes $\Hi_d$ is partitioned into the set partition $\I=\{I,J\}$, which is made by two infosets $I=\{h_1\}$ and $J=\{h_2, h_3\}$.

The sets of sequences are constructed as follows.
For the chance agent, we have that $\Sigma_c=\{ (h_0,a), (h_0, b), (h_0, c) \}$, while for the receiver we have that $\Sigma_r=\{ (I,d), (I,e), (J, f), (J,g) \}$.
Let us remark that, since the receiver cannot distinguish between nodes $h_2$ and $h_3$, it only has $2$ sequences originating from such nodes; namely $(J,f)$ and $(J,g)$.
Finally, the sender can be thought of as a perfect-information agent selecting action recommendations for the receiver at decision nodes, so that their set of sequences is $\Sigma_s=\{ (h_1,d), (h_1,e), (h_2,f), (h_2,g), (h_3,f), (h_3, g) \}$.

%% file: src/appendix/appendix_notation.tex
\section{Additional notation needed in the proofs}


In this section, we introduce some additional notation that will be useful in the proofs.

We denote by $\Pi_r\coloneqq\X_r\cap\{0,1\}^{|\Sigma_r|}$ the set of \emph{deterministic} sequence-form strategies (a.k.a. \emph{pure strategies}) of the receiver, which are the strategies specifying to play a single action with probability one at each infoset.
%
%
%
The set of receiver's deterministic strategies in the SDM sub-problem that starts from an infoset $I \in \I$ is denoted as
\(
\Pi_{r,I} \defeq \X_{r,I}\cap\{0,1\}^{|\Sigma_r|}.
\)
Moreover, we let $\Sigma_{r,I} \subseteq \Sigma_{r}$ be the set of receiver's sequences in the SDM sub-problem that starts from an infoset $I \in \I$; formally, $\Sigma_{r,I} \coloneqq \left\{  \sigma \in \Sigma_{r} \mid \sigma_r(I) \preceq \sigma \wedge \exists z \in \Z(I) : \sigma \preceq\sigma_r(z)  \right\}$ 

Given any infoset $I \in \I$, we let $\Z(I)\subset\Z$ be the set of terminal nodes $z \in \Z$ such that the path from the root of the tree to $z$ passes through a node in $I$.
Moreover, given $\sigma=(I,a)$ with $a \in A(I)$, we define $\Z(\sigma)=\Z(I,a)\subset \Z(I)$ as the set of terminal nodes whose corresponding paths include playing action $a$ at a node in $I$.
For every infoset $I \in \I$, we also introduce a function $h_I: Z(I)\to I$ such that $h_I(z)$ defines the unique node $h\in I$ on the path from the root of the tree to $z$.

Given an infoset $I \in \I$ and an action $a \in A(I)$, we define ${C}(I,a) \subseteq \I$ as the set of all the infosets which immediately follow infoset $I$ through action $a$, \emph{i.e.}, those infosets $J \in \I$ such that $\sigma_r(J) = (I,a)$.
Moreover, we let $C(I) \subseteq \I$ be the set of all infosets that follow $\I$, \emph{i.e.}, those infosets $J \in \I$ such that there exits $a \in A(I)$ with $\sigma=(I,a)$ such that $\sigma \preceq \sigma_r(J)$.

%% file: src/appendix/appendix_offline.tex
\section{Proofs omitted from Section~\ref{sec:known_prior}}\label{sec_app:proof_4}

Let us remark that all the results in Section~\ref{sec:known_prior} can be straightforwardly generalized to the case of a generic $\muvec \in \X_c$, as needed for the proofs of the results in Sections~\ref{sec:full_feedback}~and~\ref{sec:unkown_prior}.
For ease of exposition, we state and prove the results of Section~\ref{sec:known_prior} for the prior $\muvec^\star$. 

First, we prove a preliminary lemma that allows us to express the receiver's expected utility  difference between using a $(\sigma,\rhovec_\sigma)$-SPDP and following action recommendations by only considering the terminal nodes under the infoset in which the SPDP prescribed to deviate.
A similar result for the case of correlated strategies can be found in~\cite[Appendix~A]{celli2020no}.

\begin{restatable}{lem}{tmpe}
	\label{lem:decomposition_1}
	Given $\phivec \in \Phi$, for every $(\sigma, \rhovec_\sigma)$-SPDP with $\sigma=(I,a) \in \Sigma_r$ and $\rhovec_\sigma \in \X_{r, I}$, it holds:
	\begin{align*}
	\V_{\sigma\to\rhovec_\sigma}(\phivec,\muvec^\star) - \V(\phivec,\muvec^\star) = & \sum\limits_{z\in Z(I) }\phivec[(h_I(z),a)] \rhovec_\sigma[\sigma_r(z)]\muvec^\star[\sigma_c(z)]\ur(z) + \\
	& -\sum\limits_{z\in Z(\sigma)} \phivec[\sigma_s(z)]\muvec^\star[\sigma_c(z)]\ur(z).
	\end{align*}
\end{restatable}

\begin{proof}
	We define the following three disjoint events for any $(\sigma,\rhovec_\sigma)$-SPDP, where $\sigma=(I,a  )$.
	\begin{enumerate}[start=1,label={(C\arabic*):}]
		\item A terminal node $z\in\Z(\sigma)$ is reached.
		\item A terminal node $z\in\Z(I,a^\prime)$ for some $a^\prime\neq a \in A(I)$ is reached.
		\item A terminal node $z\in\Z/\Z(I)$ is reached.
	\end{enumerate}
	Next, under each event, we define the probability $p_{\sigma\to\rhovec_\sigma}(z)$ of reaching a terminal node $z$:
	\begin{enumerate}[start=1,label={(C\arabic*):}]
	\item Since $z\in\Z(\sigma)$, the node $z$ is reached by means of the continuation strategy $\rhovec_\sigma$. Thus:
	\[
		p^{(1)}_{\sigma\to\rhovec_\sigma}(z)\coloneqq \phivec[(h_I(z),a)]\muvec^\star[\sigma_c(z)]\rhovec_\sigma[\sigma_r(z)].
	\]
	\item Since $z\in\Z(I,a^\prime)$ for $a^\prime\neq a \in A(I)$, the node $z$ can be reached either by deviating and then committing to the continuation strategy $\rhovec_\sigma$ or by following recommendations.
	Moreover, these two cases are exclusive, and, thus, we can write:
	\[
		p^{(2)}_{\sigma\to\rhovec_\sigma}(z)\coloneqq
		\phivec[(h_I(z),a)]\muvec^\star[\sigma_c(z)]\rhovec_\sigma[\sigma_r(z)]+\phivec[\sigma_s(z)]\muvec^\star[\sigma_c(z)].
	\]
	\item Since $z\in\Z/\Z(I)$, the node $z$ is reached by following recommendations: 
	\[
		p^{(3)}_{\sigma\to\rhovec_\sigma}(z)\coloneqq 
		\phivec[\sigma_s(z)]\muvec^\star[\sigma_c(z)].
	\]
\end{enumerate}

We observe that $p^{(2)}_{\sigma\to\rhovec_\sigma}(z)=p^{(1)}_{\sigma\to\rhovec_\sigma}(z)+p^{(3)}_{\sigma\to\rhovec_\sigma}(z)$, and, thus, we can write $\V_{\sigma\to\rhovec_\sigma}(\phivec, \muvec^\star)$ as:
\begin{align*}
\V_{\sigma\to\rhovec_\sigma}(\phivec, \muvec^\star)&\coloneqq\sum\limits_{z\in\Z(\sigma)}p^{(1)}_{\sigma\to\rhovec_\sigma}(z)\ur(z)+\sum\limits_{\substack{z\in\Z(I,a^\prime):\\ a^\prime\neq a \in A(I)}}p^{(2)}_{\sigma\to\rhovec_\sigma}(z)\ur(z)+\sum\limits_{z\in\Z/\Z(I)}p_{\sigma\to\rhovec_\sigma}^{(3)}(z)\ur(z)\\
&\le \sum\limits_{z\in\Z(I)}p_{\sigma\to\rhovec_\sigma}^{(1)}(z)\ur(z)+\sum\limits_{z\in\Z/\Z(\sigma)}p_{\sigma\to\rhovec_\sigma}^{(3)}(z)\ur(z).
\end{align*}
Furthermore, by using the definition of $p^{(3)}_{\sigma\to\rhovec_\sigma}(z)$, we can write $\V(\phivec,\muvec)\coloneqq\sum_{z\in\Z}\ur(z)p^{(3)}_{\sigma\to\rhovec_\sigma}(z)$. Thus:
\[
\V_{\sigma\to\rhovec_\sigma}(\phivec, \muvec^\star)-\V(\phivec,\muvec^\star)=\sum\limits_{z\in\Z(I)} p_{\sigma\to\rhovec_\sigma}^{(1)}(z)\ur(z)-\sum\limits_{z\in\Z(\sigma)}p_{\sigma\to\rhovec_\sigma}^{(3)}(z)\ur(z),
\]
which is the statement of the lemma by substituting the definitions of $p_{\sigma\to\rhovec_\sigma}^{(1)}(z)$ and $p_{\sigma\to\rhovec_\sigma}^{(3)}(z)$.
\end{proof}

Now, we exploit Lemma~\ref{lem:decomposition_1} to prove the following local decomposition of a DP into SPDPs.

\decomposition*
\begin{proof}
	For any terminal node $z\in\Z$, let $p^{\omegavec\to\rhovec}(z; \phivec,\muvec^\star)$ be the probability of reaching node $z$ when the receiver employs the $(\omegavec,\rhovec)$-DP under the signaling scheme $\phivec$ and the prior $\muvec^\star$.
	It holds:
	\begin{align*}
	p^{\omegavec\to\rhovec}(z; \phivec,\muvec^\star) \coloneqq & \sum\limits_{\substack{\sigma=(I,a) \in \Sigma_r:  \sigma \preceq \sigma_r(z)}}\omegavec[\sigma] \phivec[(h_I(z),a)]\rhovec_\sigma[\sigma_r(z)]\muvec^\star[\sigma_c(z)] +
	\\
	& +\phivec[\sigma_s(z)]\muvec^\star[\sigma_c(z)]\left(1-\sum\limits_{\substack{\sigma\in\Sigma_r: \sigma\preceq \sigma_r(z)}}\omegavec[\sigma]\right).
	\end{align*}
	
	The sum in the first term in the definition of $p^{\omegavec\to\rhovec}(z; \phivec,\muvec^\star)$ accounts for the probabilities of reaching $z$ when the receiver reaches infoset $I$, is recommended to play action $a$, and deviates by following the continuation strategy $\rhovec_\sigma$ thereafter, for all the sequences $\sigma = (I,a)$ that precede the sequence $\sigma_r(z)$ reaching $z$.
	Instead, the second term in the definition of $p^{\omegavec\to\rhovec}(z; \phivec,\muvec^\star)$ accounts for the probability of reaching $z$ by following recommendations.
	Thus, $\V^{\omegavec\to\rhovec}(\phivec, \muvec^\star)=\sum_{z\in Z} p^{\omegavec\to\rhovec}(z; \phivec,\muvec^\star)u(z)$.

	By rearranging the terms in $\V^{\omegavec\to\rhovec}(\phivec, \muvec^\star)$, we get to the following result:
	\begin{align}
		\V^{\omegavec\to\rhovec}(\phivec, \muvec^\star) &= \V(\phivec, \muvec^\star) + \sum\limits_{z\in \Z}\Bigg[ \sum\limits_{\substack{\sigma=(I,a): \sigma\preceq \sigma_r(z)}}  \omegavec[\sigma] \phivec[(h_I(z),a)]\rhovec_\sigma[\sigma_r(z)]\muvec^\star[\sigma_c(z)]\ur(z) + \nonumber \\
		& \hspace{3.3cm} -\sum\limits_{\substack{\sigma \in \Sigma_r : \sigma\preceq \sigma_r(z)}} \omegavec[\sigma] \phivec[\sigma_s(z)]\muvec^\star[\sigma_c(z)]\ur(z) \Bigg]\nonumber\\
		& =\V(\phivec, \muvec^\star) -\sum\limits_{\sigma\in\Sigma_r} \omegavec[\sigma]\sum\limits_{z\in \Z(\sigma)} \phivec[\sigma_s(z)]\muvec^\star[\sigma_c(z)]\ur(z) + \nonumber\\
		& \hspace{0.5cm}+\sum\limits_{\sigma\in\Sigma_r}\omegavec[\sigma]\sum\limits_{z\in \Z(I)}  \phivec[(h_I(z),a)] \rhovec_\sigma[\sigma_r(z)]\muvec^\star[\sigma_c(z)]\ur(z).\label{eq:decomposition_1}
	\end{align}
	
	Thus, by combining Lemma~\ref{lem:decomposition_1} with Equation~\eqref{eq:decomposition_1} we get that:
	\[
		\V^{\omegavec\to\rhovec}(\phivec, \muvec^\star) - \V(\phivec,\muvec^\star) = \sum\limits_{\sigma\in\Sigma_r}\omegavec[\sigma]\left[\V_{\sigma\to\rhovec_\sigma}(\phivec,\muvec^\star)-\V(\phivec,\muvec^\star)\right],
	\]
	which concludes the proof.
\end{proof}

\corollaryone*

\begin{proof}
	By using Theorem~\ref{th:decomposition_zin}, we derive the following:
	\begin{align*}
		\max\limits_{(\omegavec,\rhovec)\in\Omega\times \mathcal{P}}\V^{\omegavec\to\rhovec}(\phivec,\muvec^\star)- \V(\phivec,\muvec^\star)&=\max\limits_{(\omegavec,\rhovec)\in\Omega\times \mathcal{P}}	\sum\limits_{\sigma\in\Sigma_r}\omegavec[\sigma]\left(\V_{\sigma\to\rhovec_\sigma}(\phivec,\muvec^\star)-\V(\phivec,\muvec^\star)\right)\\
		& \le 	\max\limits_{(\omegavec,\rhovec)\in\Omega\times \mathcal{P}}	\sum\limits_{\sigma\in\Sigma_r}\omegavec[\sigma]\left[ \V_{\sigma\to\rhovec_\sigma}(\phivec,\muvec^\star)-\V(\phivec,\muvec^\star)\right]^+\\
		&\le\max\limits_{\rhovec\in \mathcal{P}}\sum\limits_{\sigma\in\Sigma_r}\left[\V_{\sigma\to\rhovec_\sigma}(\phivec,\muvec^\star)-\V(\phivec,\muvec^\star)\right]^+\\
		&= \sum\limits_{\sigma\in\Sigma_r} \left[\max\limits_{\rhovec_\sigma\in\X_{r,I}} \V_{\sigma\to\rhovec_\sigma}(\phivec,\muvec^\star)-\V(\phivec,\muvec^\star)\right]^+.
	\end{align*}
	This concludes the proof.
\end{proof}

\polytope*

\begin{proof}
	%
	In order to prove that the set $\Lambda_\epsilon(\muvec^\star)$ can be described by means of linear constraints, we employ duality arguments related to the $\max$ problem in the definition of $\Lambda_\epsilon(\muvec^\star)$ (Definition~\ref{def:pers_poly}).
	%

	By Lemma~\ref{lem:decomposition_1}, for every sequence $\sigma=(I,a) \in \Sigma_r$, we can rewrite the expression in the left-hand side of the inequality characterizing $\Lambda_\epsilon(\muvec^\star)$ in Definition~\ref{def:pers_poly} as follows:
	\[
	\max_{\rhovec_\sigma \in \X_{r,I}} \left\{  \sum_{z\in Z(I) }\phivec[(h_I(z),a)] \rhovec_\sigma[\sigma_r(z)]\muvec^\star[\sigma_c(z)]\ur(z)\right\} -\sum_{z\in Z(\sigma)} \phivec[\sigma_s(z)]\muvec^\star[\sigma_c(z)]\ur(z),
	\]
	so that $\Lambda_\epsilon(\muvec^\star)$ can be expressed as the set of all $\phivec \in \Phi$ such that the above expression has value less than or equal to $\epsilon/|\Sigma_{r}|$ for every $\sigma \in \Sigma_{r}$.
	Observe that the expression in the $\max$ operator is a linear function of $\rhovec_\sigma$, and that the set $\X_{r,I}$ is a polytope by definition.
	Thus, for every $\sigma=(I,a) \in \Sigma_r$, the maximization above can be equivalently rewritten as the following linear program:
	\begin{subequations}\label{prob:primal_max}
		\begin{align}
			\max_{\xvec^{I,a} \geq \boldsymbol{0}} \,\, & \left( \xvec^{I,a} \right)^\top \cvec(\phivec,\muvec^\star) \quad \text{s.t.} \\
			& \Fmat_I \xvec^{I,a}  =\fvec_I \label{cons_abs_0}
		\end{align}
	\end{subequations}
where $\xvec^{I,a}$ is a vector of variables indexed over sequences $\Sigma_{r,I} \cup \{ \sigma_r(I) \}$.
%
%
Notice that $c(\phivec,\muvec^\star)\in\mathbb{R}^{|\Sigma_{r,I}|}$ is a vector of coefficients such that the component corresponding to each $\sigma' \in\Sigma_{r,I} $ is
\[
c(\phivec,\muvec^\star)[\sigma'] \coloneqq \sum\limits_{z \in \Z(I): \sigma_r(z) = \sigma'} \phivec[(h_I(z), a)]\muvec^\star[\sigma_c(z)]\ur(z),
\]
while $c(\phivec,\muvec^\star)[\sigma_r(I)] \coloneqq0$.
Moreover, $\Fmat_I\in\{-1, 0 , 1\}^{(1+|C(I)|)\times |\Sigma_{r,I}|}$ is a matrix of coefficients whose components are defined as follows: $[\Fmat_I]_{I_\varnothing, \sigma_r(I)}\coloneqq 1$ and $[\Fmat_I]_{I_\varnothing, \sigma'}\coloneqq 0$ for all sequences $\sigma' \in\Sigma_{r,I} $, where $I_\varnothing$ is a fictitious infoset indexing the first row, while, for every infoset $J\in C(I)$ following $I$ (this included) and sequence $\sigma' \in\Sigma_{r,I} \cup \{\sigma_r(I)\}$:
\begin{equation*}
	[\Fmat_I]_{J, \sigma'}\coloneqq
	\begin{cases}
		-1 & \text{if}\quad \sigma' = \sigma_r(J)\\
		1 & \text{if}\quad \sigma' = (J, a') \, \text{for some} \, a' \in A(J)\\
		0 & \text{otherwise} 
	\end{cases}.
\end{equation*}
Finally, $\fvec_I\in\{0,1\}^{1+ |C(I)| }$ is a vector whose components are all zero apart from that one corresponding to the sequence $\sigma_r(I)$, which is one (see also~\citep{koller1996efficient}).

The dual linear program of Problem~\eqref{prob:primal_max} reads as:
	\begin{subequations}\label{prob:dual_min}
	\begin{align}
		\min_{\yvec^{I,a}} \,\,  & \yvec^{I,a}[I_\varnothing] \quad \text{s.t.} \\
		& \Fmat_I^\top\yvec^{I,a} \geq  \cvec(\phivec,\muvec^\star), \label{eq:dual_constraints}
	\end{align}
\end{subequations}
where $\yvec^{I,a}$ is a vector of dual variables indexed over $C(I) \cup \{ I_\varnothing \}$.
For ease of notation, we let $\texttt{OPT}_{I,a}$ be the optimal value of Problem~\eqref{prob:dual_min} instantiated for the sequence $\sigma = (I,a)$. 

By strong duality, we have that the optimal value of the primal (Problem~\eqref{prob:primal_max}) is equal to the optimal value of the dual (Problem~\eqref{prob:dual_min}), and this allows us to readily rewrite the set $\Lambda_\epsilon(\muvec^\star)$ as follows:
\begin{align}\label{eq:polytope_min}
	\Lambda_\epsilon(\muvec^\star) = \Bigg\{\phivec\in\Phi \, \Big\vert\, & \texttt{OPT}_{I,a}-\sum_{z\in Z(\sigma)} \phivec[\sigma_s(z)]\muvec^\star[\sigma_c(z)]\ur(z)\le \frac{\epsilon}{|\Sigma_r|}\quad\forall\sigma=(I,a)\in\Sigma_r \Bigg\}.
\end{align}
Moreover, we can remove $\texttt{OPT}_{I,a}$ in Equation~\eqref{eq:polytope_min} since it appears in in the right-hand side of a $\leq$ inequality and Problem~\eqref{prob:dual_min} is a $\min$ problem.
Thus, the set $\Lambda_\epsilon(\muvec^\star)$ can be written as follows:
\begin{align}\label{eq:polytope_min_2}
	\Lambda_\epsilon(\muvec^\star) = \Bigg\{\phivec\in\Phi\, \Big\vert\, & \exists\yvec^{I,a} \in \mathbb{R}^{1+|C(I)|}: \yvec^{I,a}[I_\varnothing]-\sum_{z\in Z(\sigma)} \phivec[\sigma_s(z)]\muvec^\star[\sigma_c(z)]\ur(z)\le \frac{\epsilon}{|\Sigma_r|} \nonumber \\
	& \wedge \Fmat_I^\top\yvec^{I,a} \geq  \cvec(\phivec,\muvec^\star)\quad\forall\sigma=(I,a)\in\Sigma_r \Bigg\},
\end{align}
which is comprised of a polynomial number of inequalities and variables, concluding the proof.

Let us also notice that, by expanding the constraints of Problem~\eqref{prob:dual_min}, one can easily check that they can be equivalently rewritten recursively, as follows.
For every sequence $\sigma'=(J,a') \in\Sigma_{r,I}$, Constraints~\eqref{eq:dual_constraints} can be rewritten as:
%
\begin{equation}\label{eq:recursive}
	\yvec^{I,a}[J]\ge \sum\limits_{z \in \Z(I):  \sigma_r(z)=(J,a')} \phivec[(h_I(z), a)]\muvec^\star[\sigma_c(z)]\ur(z)+\sum\limits_{K\in C{(J,a')}} \yvec^{I,a}[K],
\end{equation}
while, for sequence $\sigma_r(I)$, Constraint~\eqref{eq:dual_constraints} can be written as
\(
	\yvec^{I,a}[I_\varnothing]\ge\yvec^{I,a}[I].
\)
%
Intuitively, at any optimal solution to Problem~\eqref{prob:dual_min}, we can interpret the value of the dual variable $\yvec^{I,a}[I_\varnothing]$ as the receiver's expected utility obtained by playing the best possible continuation strategy after being recommended action $a$ at infoset $I$.
Indeed, the first term in the right-hand-side of Equation~\eqref{eq:recursive} is the utility immediately obtainable after playing $a'$ at infoset $J$, while the second term recursively encodes the utilities obtained (non-immediately) following $a'$ at $J$.  
%
%
%
%
\end{proof}

\containment*
\begin{proof}
	First, we prove that $\pers(\muvec^\star)\equiv\Lambda(\muvec^\star)$.
	Suppose that $\phivec \in \pers(\muvec^\star)$. Then,  Definition~\ref{def:persuasiveness} implies \[
	\V^{\omegavec\to\rhovec}(\phivec,\muvec^\star)- \V(\phivec,\muvec^\star)\le 0,
	\]
	for every $\omegavec\in\Omega$ and $\rhovec\in \mathcal{P}$.
	Thus, by Theorem~\ref{th:decomposition_zin} we have that:
	\[
	\sum\limits_{\sigma\in\Sigma_r}\omegavec[\sigma] \Big(\V_{\sigma\to\rhovec_\sigma}(\phivec,\muvec^\star) - \V(\phivec,\muvec^\star)\Big)\le 0,
	\]
	for every $\omegavec\in\Omega$ and $\rhovec\in \mathcal{P}$, which implies that:
		\[
	\max\limits_{\rhovec_\sigma \in \X_{r,I}} \V_{\sigma\to \rhovec_\sigma}(\phivec,\muvec^\star)-\V(\phivec,\muvec^\star)\le 0\quad \forall \sigma\in\Sigma_r,
	\]
	and $\phivec \in \Lambda(\muvec^\star)$, proving the first part of the statement.

	On the other hand, $\Lambda(\muvec^\star)\subseteq \pers(\muvec^\star)$ is directly implied by Corollary~\ref{cor:decompostion_regret}. Thus, $\Lambda(\muvec^\star)\equiv\pers(\muvec^\star)$.
%
%
Moreover, from Definition~\ref{def:pers_poly} it trivially follows that  $\Lambda(\muvec^\star)\subseteq \Lambda_\epsilon(\muvec^\star)$.

Finally, we prove that $\Lambda_\epsilon(\muvec^\star)\subseteq \pers_\epsilon (\muvec^\star)$.
Given $\epsilon>0$, let $\phivec \in \Lambda_\epsilon(\muvec^\star)$.
By Corollary~\ref{cor:decompostion_regret}, it holds:
\begin{align*}
	\max\limits_{(\omegavec,\rhovec)\in\Omega\times \mathcal{P}}\V^{\omegavec\to\rhovec}(\phivec,\muvec^\star)- \V(\phivec,\muvec^\star)& \le \sum\limits_{\sigma= (I,a)\in\Sigma_r} \left[\max\limits_{\rhovec_\sigma \in\X_{r,I}} \V_{\sigma\to\rhovec_\sigma}(\phivec,\muvec^\star)-\V(\phivec,\muvec^\star)\right]^+\\
	& \le \sum\limits_{\sigma= (I,a)\in\Sigma_r} \frac{\epsilon}{|\Sigma_r|}= \epsilon,
\end{align*}
which implies that $\phivec\in \pers_\epsilon(\muvec^\star)$.
This concludes the proof.
\end{proof}

\offline*

\begin{proof}
	It easy to check that the problem can be written as the following linear program:
	\[
	\max_{\phivec \in \Lambda(\muvec^\star)} \U(\phivec,\muvec^\star),
	\]
	where the objective function is linear and $\Lambda(\muvec^\star)$ is a polytope that can be represented by a polynomial number of linear inequalities, by Lemma~\ref{thm:polytope}.
\end{proof}

%% file: src/appendix/appendix_online.tex
\section{Proofs omitted from Section~\ref{sec:online}}\label{sec_app:proof_5}

\impossibility*

\begin{figure}[!ht]
	\centering
	\includegraphics[width=0.27\textwidth]{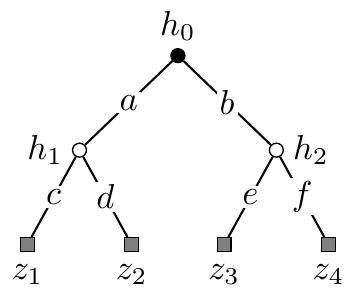}
	\caption{Tree structure used in the proof of Theorem~\ref{th:impossibility}. Black round nodes are decision nodes $\Hi_d$. White round nodes are the chance nodes $\Hi_c$, while grey square nodes are the terminal nodes $\Z$.}
	\label{fig:tree_impossibility}
\end{figure}

\begin{proof}
	We define two instances $\inst{i}$ and $\inst{j}$ of BPSDM problem based on the tree structure presented in Figure~\ref{fig:tree_impossibility}.
	In instance $\inst{i}$, respectively $\inst{j}$, the prior is defined as follows:
	\[
	\inst{i}\coloneqq
	\begin{cases}
		\muvec^\star[(h_1, c)] = \frac{1}{2}+\epsilon\\
		\muvec^\star[(h_1, d)] = \frac{1}{2}-\epsilon\\
		\muvec^\star[(h_2, e)] = \frac{1}{2}-\epsilon\\
		\muvec^\star[(h_2, f)] = \frac{1}{2}+\epsilon\\
	\end{cases},
	\]
	\[
	\inst{j}\coloneqq
	\begin{cases}
		\muvec^\star[(h_1, c)] = \frac{1}{2}-\epsilon\\
		\muvec^\star[(h_1, d)] = \frac{1}{2}+\epsilon\\
		\muvec^\star[(h_2, e)] = \frac{1}{2}+\epsilon\\
		\muvec^\star[(h_2, f)] = \frac{1}{2}-\epsilon\\
	\end{cases}.
	\]
	Moreover, for both instances $\ur(z_1)=\ur(z_3)=1$ and  $\ur(z_2)=\ur(z_4)=0$.
	A direct computation shows that, in instance $\inst{i}$, it holds $\Rr_T^{\inst{i}}=2\epsilon \sum_{t=1}^T\phivec_t[(h_0, b)]$, while one can similarly compute that $\Rr_T^{\inst{j}}=2\epsilon \sum_{t=1}^T\phivec_t[(h_0, a)]$.
	Let $\mathbb{P}^{\inst{i}}$ and $\mathbb{P}^{\inst{j}}$ be the probability measures of instance $\inst{i}$ and $\inst{j}$, respectively.
	Assume that $\mathbb{P}^{\inst{j}}[\Rr_T^{\inst{j}}\le 0]\ge 1-\delta$.
	Then, we know from the Pinsker inequality that:
	\[
	\mathbb{P}^\inst{i}\left[\sum\limits_{t=1}^T\phivec_t[(h_0, a)]\le 0\right]\ge 1-\sqrt{\frac{1}{2}\mathcal{K}(\inst{i},\inst{j})}-\delta,
	\]
	where $\mathcal{K}(\inst{i},\inst{j})$ is the Kullback-Leibler divergence between instance $\inst{i}$ and $\inst{j}$.
	By using the Kullback-Leibler decomposition (see, \emph{e.g.}, \cite{lattimore2020bandit} for more details), we can state that:
	\[
	\mathcal{K}(\inst{i},\inst{j})=2T\mathcal{K}(B_{1/2+\epsilon}, B_{1/2-\epsilon}),
	\]
	where $\mathcal{K}(B_{1/2+\epsilon}, B_{1/2-\epsilon})\le16\epsilon^2$ is the Kullback-Leibler divergence between a Bernoulli of parameter $1/2+\epsilon$ and one of parameter $1/2-\epsilon$.
	Thus:
	\[
	\mathbb{P}^\inst{i}\left[\sum\limits_{t=1}^T\phivec_t[(h_0, a)]\le 0\right]\ge 1-4\epsilon\sqrt{T}-\delta.
	\]
	Moreover, in instance $\inst{i}$, we have that $\Rr_T^{\inst{i}}=2\epsilon\sum_{t=1}^T\phivec_t[(h_0, b)]$, which implies:
	\[
	\mathbb{P}^\inst{i}\left[ \Rr_T^{\inst{i}}\ge 2\epsilon T\right]\ge 1-4\epsilon\sqrt{T}-\delta.
	\]
	By setting $\epsilon=\frac{1}{16\sqrt{T}}$, we have that:
	\[
	\mathbb{P}^\inst{i}\left[ \Rr_T^{\inst{i}}\ge \frac{1}{8}\sqrt{T}\right]\ge 0.75-\delta.
	\]
	Thus, any algorithm that guarantees with high probability $R_T^r\le 0$ in instance $\inst{j}$ fails with high probability in instance $\inst{i}$.
	This proves the claim.
\end{proof}

%% file: src/appendix/appendix_full_feedback.tex
\section{Proofs omitted from Section~\ref{sec:full_feedback}}\label{app:full_feed}

Before presenting the proofs of the results in Section~\ref{sec:full_feedback} ,we introduce some preliminary lemmas.

\begin{restatable}{lem}{tmpq}\label{lem:persuasiveness_derivative}
	Given any $\phivec\in\Phi$ and $\muvec, \muvec' \in \X_c$, if it is the case that $\phivec\in\Lambda_\epsilon(\muvec)$ and $\|\muvec-\muvec^\prime\|_\infty\le\gamma$, then it holds that $\phivec\in\Lambda_{\epsilon^\prime}(\muvec^\prime)$ with $\epsilon^\prime=2|\Z||\Sigma_r|\gamma+\epsilon$.
\end{restatable}
\begin{proof}
	For every $(\sigma,\rhovec_\sigma)$-SPDP with $\sigma=(I,a)$, the following inequalities hold:
	\begin{align}
		\V_{\sigma\to\rhovec_\sigma}(\phivec,\muvec^\prime)-&\V(\phivec,\muvec^\prime)\nonumber \\
		=&\sum\limits_{\Z(I)} \phivec[(h_I(z),a)]\rhovec_\sigma[\sigma_r(z)]\muvec^\prime[\sigma_c(z)]\ur(z)-\sum\limits_{z\in Z(\sigma)}\phivec[\sigma_s(z)]\muvec^\prime[\sigma_c(z)]\ur(z) \nonumber\\
		\le&\sum\limits_{\Z(I)} \phivec[(h_I(z),a)]\rhovec_\sigma[\sigma_r(z)]\left( \muvec^\prime[\sigma_c(z)]-\muvec[\sigma_c(z)] \right)\ur(z)\nonumber \\
		&-\sum\limits_{z\in \Z(\sigma)}\phivec[\sigma_s(z)] \left( \muvec^\prime[\sigma_c(z)]-\muvec[\sigma_c(z)] \right)\ur(z)+\frac{\epsilon}{|\Sigma_r|}\nonumber \\
		\le& 2|\Z|\|\muvec-\muvec^\prime \|_{\infty}+\frac{\epsilon}{|\Sigma_r|}\le 2|\Z|\gamma+\frac{\epsilon}{|\Sigma_r|},\nonumber
	\end{align}
	where in the first inequality we added and subtracted the difference $\V_{\sigma\to\rhovec_\sigma}(\phivec,\muvec)-\V(\phivec,\muvec)$ and used the fact that $\phivec\in\Lambda_\epsilon(\muvec)$, while the second-to-last inequality follows from H\"older's inequality. 
	Since $\V_{\sigma\to\rhovec_\sigma}(\phivec,\muvec^\prime)-\V(\phivec,\muvec^\prime)\le2|\Z|\gamma+\frac{\epsilon}{|\Sigma_r|}\coloneqq\frac{\epsilon^\prime}{|\Sigma_r|}$ holds for every $(\sigma,\rhovec_\sigma)$-SPDP, we have that $\phivec\in\Lambda_{\epsilon^\prime}(\muvec^\prime)$ with $\epsilon^\prime=|\Z||\Sigma_r|\gamma+\epsilon$, concluding the proof.
\end{proof}

\begin{restatable}{lem}{tmpw}\label{lem:highprob}
	Given any $\delta \in (0,1)$, Algorithm~\ref{alg:full_feedback} guarantees that $\mathbb{P}[\mathcal{E}]\ge1-\delta$, where:
	\[
	\mathcal{E}\coloneqq\left\{\|\widehat\muvec_t-\muvec^\star\|_\infty\le\epsilon_t \quad \forall t\in[T]\right\},
	\]
	and $\epsilon_{t}$ is chosen according to Algorithm~\ref{alg:full_feedback}.
\end{restatable}

\begin{proof}
	Let $\B_t(\delta)$ be defined as follows:
	\[
	\B_t(\delta) \coloneqq\left\{\muvec  \, \Big\vert\, \lvert\muvec[\sigma]-\widehat\muvec_{t}[\sigma]\rvert \le\sqrt{\frac{\log(2T|\Sigma_c|/\delta)}{2t}} \,\,\, \forall\sigma\in\Sigma_c\right\}.
	\]
	Clearly, $\mathbb{P}[\mathcal{E}]=\mathbb{P}[\muvec^\star\in\B_t(\delta)\, \forall t\in[T]]$.
	By Hoeffding’s inequality, we have that:
	\[
	\mathbb{P}\left(\lvert\muvec^\star[\sigma]-\widehat\muvec_{t}[\sigma]\rvert\le\sqrt{\frac{\log(2T|\Sigma_c|/\delta)}{2t}} \right)\ge 1-\frac{\delta}{T |\Sigma_c|}.
	\]
	By a union bound over $\sigma\in\Sigma_c$ and $t\in[T]$, we get that:
	\[
	\mathbb{P}\left(\lvert\muvec^\star[\sigma]-\widehat\muvec_{t}[\sigma]\rvert\le\sqrt{\frac{\log(2T|\Sigma_c|/\delta)}{2t}} \,\,\, \forall\sigma\in\Sigma_c \,\, \forall t\in[T] \right)\ge 1-{\delta}.
	\]
	This concludes the proof of the lemma.
\end{proof}

\begin{restatable}{lem}{lemmaduefull}\label{lem:lemma1}
	If the event $\mathcal{E}$ holds, Algorithm~\ref{alg:full_feedback} guarantees that $\phivec^\star\in \Lambda_{\beta_t}(\widehat\muvec_t)$ for all $t \in [T]$.
\end{restatable}

\begin{proof}
	By definition, we have that $\phivec^\star\in\Lambda(\muvec^\star)$. Moreover, since we conditioned on $\mathcal{E}$, we have that:
	\[
	\|\muvec^\star-\widehat\muvec_t\|_\infty\le\epsilon_t \,\,\,\forall t\in[T].
	\]
	Thus, we can exploit Lemma~\ref{lem:persuasiveness_derivative}, which, by letting $\beta_t \coloneqq 2|\Z||\Sigma_r|\epsilon_t$, gives that $\phivec^\star\in\Lambda_{\beta_t}(\widehat\muvec_t)$.
\end{proof}

\begin{restatable}{lem}{lemmaunofull}\label{lem:lemmaunofull}
	If the event $\mathcal{E}$ holds, Algorithm~\ref{alg:full_feedback} guarantees
	%
	that $\phivec_t \in \Lambda_{2\beta_t} (\muvec^\star)$ for all $t \in [T]$.
\end{restatable}
\begin{proof}
	Given how Algorithm~\ref{alg:full_feedback} works, we have that $\phivec_t\in \Lambda_{\beta_t}(\widehat\muvec_t)$.
	On the other hand, since we conditioned on the event $\mathcal{E}$, it must be the case that $\|\muvec^\star-\widehat\muvec_t\|\le\epsilon_t$ for all $ t\in[T]$. Thus, by Lemma~\ref{lem:persuasiveness_derivative} we obtain that $\phivec_t\in\Lambda_{2\beta_t}(\muvec^\star)$, where $\beta_t$ is defined as in the proof of Lemma~\ref{lem:lemma1}.
\end{proof}

%
%
%

\onlineAlg*
\begin{proof}
	First, we bound the computational complexity of the algorithm, then we separately analyze the sender's regret $\Rs_T$ and the receiver's regret $\Rr_T$. 
	
	\paragraph{Complexity.}
	With an argument analogous to the one used for the proof of Theorem~\ref{thm:offline}, we have that the optimization problem solved by \textsc{SelectStrategy}$()$ in Algorithm~\ref{alg:full_feedback} is a polynomially-sized linear problem (Lemma~\ref{thm:polytope}).
	Hence, it can be solved in polynomial time.

%
	
	\paragraph{Sender's regret.}
	 If the event $\mathcal{E}$ holds, which happens with probability at least $1-\delta$, then: 
	 \[
	 \muvec^\star[\sigma]-\epsilon_t\le \widehat \muvec_t[\sigma]\le \muvec^\star[\sigma]+\epsilon_t,
	 \]
	 for every sequence $\sigma\in\Sigma_c$ and round $t \in [T]$.
	 This implies that, for every $\phivec\in\Phi$, we have:
	\begin{equation*}
		\U(\phivec, \muvec^\star)-|\Z|\epsilon_t \le \U(\phivec, \widehat \muvec_t)\le \U(\phivec, \muvec^\star)+|\Z|\epsilon_t.
	\end{equation*}
 	Moreover, under the event $\mathcal{E}$, we have that $\phivec^\star\in \Lambda_{\beta_t}(\widehat\muvec_t)$ and, thus, $\U(\phivec^\star,\widehat \muvec_t)\le \U(\phivec_t,\widehat \muvec_t)$ as $\phivec_t$ is computed by optimizing $\U(\cdot, \widehat \muvec_t)$ over $\Lambda_{\beta_t}(\widehat\muvec_t)$.
	%
	By putting all the above results together,
	we get that, under event $\mathcal{E}$, the following holds:
	\[
	\U(\phivec^\star, \muvec^\star)\le \U(\phivec^\star, \widehat\muvec_t) +| \Z|\epsilon_{t}\le \U(\phivec_t, \widehat \muvec_t) +| \Z|\epsilon_{t} \le \U(\phivec_t, \muvec^\star) + 2|\Z|\epsilon_t.
	\]
	By rearranging the terms, taking the sum over $t \in [T]$, and using $\sum_{t=1}^T\frac{1}{\sqrt{t}}\le 2\sqrt{T}$, we get:
	\[
	\Rs_T \coloneqq \sum_{t=1}^T \Big( \U(\phivec^\star, \muvec^\star)-\U(\phivec_t, \muvec^\star) \Big) \le2|\Z|\sum\limits_{t=1}^T\epsilon_t\le2|\Z|\sqrt{2\log(2T|\Sigma_c|/\delta)T},
	\]
	which holds under the event $\mathcal{E}$, and, thus, with probability at least $1-\delta$.

	\paragraph{Receiver's regret.}
	%
	If the event $\mathcal{E}$ holds, thanks to Lemma~\ref{lem:lemmaunofull} we have that $\phivec_t\in\Lambda_{2\beta_t}(\muvec^\star)$.
	Thus, by using Lemma~\ref{thm:containment}, we can conclude that $\phivec_t\in\pers_{2\beta_t}(\muvec^\star)$. This implies that, with probability at least $1-\delta$, the following holds:
	\[
		\Rr_T\le 2\sum\limits_{t=1}^T\beta_t\le4|\Sigma_r||\Z|\sqrt{2\log(2T|\Sigma_c|/\delta)T},
	\]
	which concludes the proof.
\end{proof}

%% file: src/appendix/appendix_bandit_feedback.tex
\section{Proofs omitted from Section~\ref{sec:unkown_prior}}\label{sec_app:proof_7}

\lemmaunbiased*

\begin{proof}
	For any signaling scheme $\phivec \in \Phi$, we have that the probability of reaching any node $h\in\Hi_c$ during a round $t < N$ (or, equivalently, that $\pvec_t[\sigma]=1$ for some chance sequence $\sigma=(h,a)$) is a Bernoulli with parameter $\muvec^\star[\sigma]\phivec[\sigma_s(h)]$.
	Thus:
	\begin{equation*}
		\mathbb{E}[\pvec_t[\sigma]]=\phivec_t[\sigma_s(h)]\muvec^\star[\sigma].
	\end{equation*}
	If we consider any deterministic signaling scheme $\pivec \in \Pi$ and a chance sequence $\sigma=(h,a)\in\Sigma_\downarrow(\pivec)$, we have that $\phivec_t[\sigma_s(h)]=1$, and, thus, the above equation simplifies to:
	\begin{equation*}
		\mathbb{E}[\pvec_t[\sigma]]=\muvec^\star[\sigma],
	\end{equation*}
which concludes the proof.
\end{proof}

\begin{restatable}{lem}{tmpt}\label{lem:azuma_hoeff}
	Given any $\delta \in (0,1)$, Algorithm~\ref{alg:bandit_feedback} guarantees that with probability at least $1-{\delta}/{2}$:
	\[
	\sum\limits_{t=N+1}^T\sum\limits_{z\in\Z}\epsilon_t[\sigma_c(z)] \phivec_t[\sigma_s(z)]\le\sqrt{\log(4T|\Sigma_c|/\delta)|\Sigma_c|T}+|\Z|\sqrt{\log (2/\delta)T},
	\]
	where the terms $\epsilon_t[\sigma]$ for $\sigma \in \Sigma_c$ and $t \in [T]$ are defined according to Algorithm~\ref{alg:bandit_feedback}.
	%
\end{restatable}

\begin{proof}
	First, let us consider the deterministic signaling scheme $\pivec_t \in \Pi$ sampled by the algorithm according to $\phivec_t$ at round $t 	\in[T]$.
	For convenience, in the following we report the definition of $\epsilon_t[\sigma]$ (according to Algorithm~\ref{alg:bandit_feedback}) for each $\sigma \in \Sigma_c$ and $t \in [T]$:
	\[
	\epsilon_t[\sigma] \coloneqq \sqrt{\frac{\log(4T|\Sigma_c|/\delta)}{2C_t[\sigma]}},
	\]
	where $C_t[\sigma]$ represents the number of rounds $t' \leq t$ in which it is the case that $\sigma \in \Sigma_\downarrow(\pivec_{t'})$.
	Then, the following chain of inequalities holds:
	\begin{subequations}
		\begin{align}
		\sum\limits_{t=N+1}^T\sum\limits_{z\in\Z} &\epsilon_t[\sigma_c(z)] \pivec_t[\sigma_s(z)]\label{eq:concentr_1}\\
		&=\sum\limits_{t=N+1}^T\sum\limits_{\substack{\sigma\in\Sigma_c: \\ \exists z\in\Z : \sigma=\sigma_c(z)}} \left(\epsilon_t[\sigma]  \sum_{\substack{\sigma' \in \Sigma_s: \\\exists z \in \Z : \sigma=\sigma_c(z) \wedge \sigma'=\sigma_s(z) }}\pivec_t[\sigma']\right)\label{eq:concentr_5}\\
		&\le \sum\limits_{t=N+1}^T\sum\limits_{\sigma=(h,a)\in\Sigma_c} \epsilon_t[\sigma]  \pivec_t[\sigma_s(h)]\label{eq:concentr_6}\\
		&= \sum\limits_{\sigma=(h,a)\in\Sigma_c}\sum\limits_{\substack{t \in [T]: \\t \geq N+1 \wedge \pivec_t[\sigma_s(h)]=1}}\epsilon_t[\sigma]\label{eq:concentr_2} \\
		&=\sum\limits_{\sigma\in\Sigma_c}\sum\limits_{t=C_{N+1}[\sigma]}^{C_T[\sigma]}\sqrt{\frac{\log(4T|\Sigma_c|/\delta)}{2t}}\label{eq:concentr_3}\\
		&\le\sum\limits_{\sigma\in\Sigma_c}\sqrt{\log(4T|\Sigma_c|/\delta)C_T[\sigma]}\label{eq:concentr_4}\\
		&\le \sqrt{\log(4T|\Sigma_c|/\delta)|\Sigma_c|T}\label{eq:concentr_7},
		\end{align}
	\end{subequations}
	
	where Equation~\eqref{eq:concentr_6} follows by the definition of sequence-form signaling scheme of the sender, Equation~\eqref{eq:concentr_2} follows by exchanging the sums over $\sigma\in\Sigma_c$ and $t\in[T]$ and recalling that $\pivec_t$ is a deterministic signaling scheme, Equation~\eqref{eq:concentr_3} holds by definition of $\epsilon$, while Equation~\eqref{eq:concentr_4} comes from $\sum_{t=1}^T \frac{1}{\sqrt{t}}\le 2\sqrt{T}$.
	Finally, Equation~\eqref{eq:concentr_7} follows from the Cauchy-Schwarz inequality.
	
	Next, we provide a similar bound on $\sum_{t=N+1}^T\sum_{z\in\Z}\epsilon_t[\sigma_c(z)] \phivec_t[\sigma_s(z)]$. We do this by exploiting the Azuma-Hoeffding inequality~\citep{cesa2006prediction}. Indeed, we have that $\mathbb{E}[\pivec_t[\sigma]\vert \mathcal{F}_{t-1}]=\phivec_t[\sigma]$, where $\mathcal{F}_{t-1}$ is the filtration generated up to time $t-1$ from the interaction between the algorithm and the BPSDM problem.
	Thus, with probability at least $1-{\delta}/{2}$ the following holds:
	\[
	\sum\limits_{t=N+1}^T\sum\limits_{z\in\Z}\epsilon_t[\sigma_c(z)] \phivec_t[\sigma_s(z)]\le \sum\limits_{t=N+1}^T\sum\limits_{z\in\Z}\epsilon_t[\sigma_c(z)]\pivec_t[\sigma_c(z)]+|\Z|\sqrt{\log (2/\delta)T}.
	\]
	By combining the equation above with Equation~\eqref{eq:concentr_4}, we obtain:
	\[
	\sum\limits_{t=N+1}^T\sum\limits_{z\in\Z}\epsilon_t[\sigma_c(z)] \phivec_t[\sigma_s(z)]\le\sqrt{\log(4T|\Sigma_c|/\delta)|\Sigma_c|T}+|\Z|\sqrt{\log (2/\delta)T}.
	\]
	This concludes the proof.
\end{proof}

\lemmaunobandit*

\begin{proof}
	The proof is similar to the one of Lemma~\ref{lem:lemmaunofull}.
	%
	%
	If the event $\tilde{\mathcal{E}}$ holds, then we have that:
	\[
	\|\muvec^\star-\widehat\muvec_N\|_\infty\le\max\limits_{\sigma\in\Sigma_c}\epsilon_t[\sigma]\coloneqq\epsilon_N.
	\]
	Moreover, $\phivec_t\in\Lambda_{\beta_N}(\widehat\muvec_N)$ and we can use Lemma~\ref{lem:persuasiveness_derivative} to conclude that $\phivec_t\in\Lambda_{\beta_N+2\epsilon_N|\Sigma_r||\Z|}(\muvec^\star)$ for all $t> N$.
	The proof follows from $\beta_N\ge2\epsilon_N|\Z||\Sigma_r|$, since $\epsilon_N\le \sqrt{\frac{\log(4T|\Sigma_c|/\delta)|\Sigma_c|}{2N}}$.
\end{proof}

\lemmaduebandit*

\begin{proof}
	
	Since $\phivec^\star\in\Lambda(\muvec^\star)$ and, under the event $\tilde{\mathcal{E}}$, it holds that:
	\[
		\|\muvec^\star-\widehat\muvec_N\|_\infty\le\max\limits_{\sigma\in\Sigma_c}\epsilon_t[\sigma]\coloneqq\epsilon_N,
	\]
	we can use Lemma~\ref{lem:persuasiveness_derivative} to conclude that $\phivec^\star\in\Lambda_{2|\Sigma_c||\Z|\epsilon_N}(\widehat\muvec_N)$.
	The proof of the first statement is concluded by observing that $\beta_N\ge2|\Sigma_r||\Z|\epsilon_N$, since $\epsilon_N\le \sqrt{\frac{\log(4T|\Sigma_c|/\delta)|\Sigma_c|}{2N}}$.
	The second statement directly follows from the observation that, under the event $\tilde{\mathcal{E}}$, it holds $\muvec^\star\in \C_t(\delta)$.
\end{proof}

\onlineAlgBandit*

\begin{proof}
	First, we bound the computational complexity of the algorithm, then we separately analyze the sender's regret $\Rs_T$ and the receiver's regret $\Rr_T$.
	
	\paragraph{Complexity.}
	First, observe that $\U(\phivec,\muvec)$ is a linear function in $\muvec$ and it only has positive terms.
	Thus, for every $\phivec\in\Phi$, the maximum over $\C_t(\delta)$ in the optimization problem solved during the second phase of the \textsc{SelectStrategy}$()$ procedure is reached on the boundary of $\C_t(\delta)$, so that larger entries of $\muvec$ provide larger objective values.
	Formally, we define:
	\[
	\muvec_t \in \arg\max\limits_{\muvec\in\C_t(\delta)} \U(\phivec,\muvec),
	\]
	which is independent of $\phivec$.
	Then, for every $\sigma\in\Sigma_c$, we have that $\muvec_t[\sigma]=\widehat\muvec_t[\sigma]+\epsilon_t[\sigma]$.
	Thus, we can compute the signaling scheme $\phivec_t$ with a linear program as follows:
	\begin{equation}\label{eq:opt_bandit}
	\phivec_t \gets \max\limits_{\phivec\in \Lambda_{\beta_t}(\widehat\muvec_t)} \U(\phivec, \muvec_t),
	\end{equation}
	and, similarly to the proof of Theorem~\ref{th:regret_full}, we have that the optimization problem in Equation~\eqref{eq:opt_bandit} is a polynomially-sized linear program by Lemma~\ref{thm:polytope}.
	Hence, it can be solved in polynomial time.

	\paragraph{Sender's regret.}
	Under the event $\tilde{\mathcal{E}}$, which happens with probability at least $1-\delta/2$, we have that $|\muvec^\star[\sigma]-\widehat\muvec_t[\sigma]|\le\epsilon_{t}[\sigma]$ for all $t> N$. Thus,
	\begin{equation}\label{eq:bandit_rtegret_1}
		\|\muvec^\star-\widehat\muvec_t\|_{\infty}\le\max\limits_{\sigma\in\Sigma_c}\epsilon_t[\sigma]\coloneqq\epsilon_N.
	\end{equation}
	
	%
	Then, we can conclude that, under event $\tilde{\mathcal{E}}$, it holds $\muvec^\star[\sigma]+2\epsilon_{t}[\sigma]\ge\muvec_t[\sigma]$. This in turn implies:
	\[
	\U(\phivec_t,\muvec_t)\le \U(\phivec_t,\muvec^\star)+2\sum\limits_{z\in\Z}\epsilon_t[\sigma_c(z)]\phivec_t[\sigma_s(z)].
	\]
	%
%
	By Lemma~\ref{lem:lemmaduebandit}, we have that, under event $\tilde{\mathcal{E}}$, it holds
	\(
	\phivec^\star\in\Lambda_{\beta_N}(\widehat\muvec_N).
	\)
	Hence, $\U(\phivec^\star,\muvec_t)\le \U(\phivec_t,\muvec_t)$ as $\phivec_t$ is computed as the optimum over $\Lambda_{\beta_N}(\muvec_t)$.
	Moreover, by Lemma~\ref{lem:lemmaduebandit} we also have that $\U(\phivec^\star,\muvec^\star)\le \U(\phivec^\star,\muvec_t)$, which implies:
	\begin{align*}
	\U(\phivec^\star,\muvec^\star)\le \U(\phivec^\star,\muvec_t)
 	\le  \U(\phivec_t,\muvec_t) 
	 \le \U(\phivec_t, \muvec^\star) + 2\sum\limits_{z\in\Z}\epsilon_t[\sigma_c(z)]\phivec_t[\sigma_s(z)].
	\end{align*}
	Then, we can decompose the sender's regret as:
	\begin{align*}
	\Rs_T&=\sum\limits_{t=1}^N\Big( \U(\phivec^\star,\muvec^\star)-\U(\phivec_t, \muvec^\star) \Big)+\sum\limits_{t=N+1}^T \Big( \U(\phivec^\star,\muvec^\star)-\U(\phivec_t, \muvec^\star) \Big)\\
	&\le N+2\sum\limits_{t=N+1}^T\sum\limits_{z\in\Z}\epsilon_t[\sigma_c(z)]\phivec_t[\sigma_s(z)].
	\end{align*}
	By using Lemma~\ref{lem:azuma_hoeff} and a union bound, we can conclude that with probability at least $1-\delta$:
	\[
	\Rs_T\le N+2\left(\sqrt{\log(4T|\Sigma_c|/\delta)|\Sigma_c|T}+|\Z|\sqrt{\log (2/\delta)T}\right).
	\]

	\paragraph{Receiver's regret.}
	By Lemma~\ref{lem:lemmaunobandit}, under the event $\tilde{\mathcal{E}}$, we have that $\phivec_t\in\Lambda_{2\beta_N}(\muvec^\star)$ for all $t\ge N$.
	Moreover, by Lemma \ref{thm:containment}, it holds that $\Lambda_{2\beta_N}(\muvec^\star) \subseteq \pers_{2\beta_N} (\muvec^\star)$.
	%
	Hence, with probability at least $1-\delta$:
	\[
	\Rr_T\le N + 2T\beta_N=N+4T|\Z||\Sigma_r|\sqrt{\frac{|\Sigma_c|\log(4T|\Sigma_c|/\delta)}{2N}}.
	\]
	This concludes the proof.
\end{proof}

\lowerbound*

\begin{proof}
	We define two instances $\inst{i}$ and $\inst{j}$ of a BPSDM problem whose tree structures are as in Figure~\ref{fig:tree_impossibility}.
	In both instances, we have that $\us(z_1)=\us(z_2)=0$ and  $\us(z_3)=\us(z_4)=1$ for the sender, while $\ur(z_1)=\ur(z_3)=1$ and $\ur(z_2)=\ur(z_4)=0$ for the receiver.
	Moreover, in both instances we have that for the chance node $h_1$ it holds $\muvec^\star[(h_1, c)]=\muvec^\star[(h_1, d)]=1/2$.
	Instead, the two instances differ in the probabilities of chance node $h_2$, which are defined as follows:
		\[
		\inst{i}\coloneqq
		\begin{cases}
			\muvec^\star[(h_2, e)] = \frac{1}{2}-\epsilon\\
			\muvec^\star[(h_2, f)] = \frac{1}{2}+\epsilon\\
		\end{cases},
		\]
		\[
		\inst{j}\coloneqq
		\begin{cases}
			\muvec^\star[(h_2, e)] = \frac{1}{2}+\epsilon\\
			\muvec^\star[(h_2, f)] = \frac{1}{2}-\epsilon\\
		\end{cases}.
		\]
	Simple calculations show that, in instance $\mathfrak{j}$, we have that the regret of the sender is:
	\[
	\Rs_T^{\inst{j}}=\sum\limits_{t=1}^T\phivec_t[(h_0, a)]
	\]
	Hence, if we require that (in high probability with respect to the measure $\mathbb{P}^{\inst{j}}$ of instance $\inst{j}$) the sender's regret is smaller than a threshold $K$, then:
	\[
	\mathbb{P}^\inst{j}\left[\sum\limits_{t=1}^T\phivec_t[(h_1, a)]\le K\right]\ge 1-\delta.
	\]
%
%

	The Pinsker's inequality states that:
	\[
	\mathbb{P}^\inst{i}\left[\sum\limits_{t=1}^T\phivec_t[(h_1, a)]\le K\right]\ge 1-\delta-\sqrt{\frac{1}{2}\mathcal{K}(\inst{j},\inst{i})},
	\]
	where $\mathcal{K}(\inst{j}, \inst{i})$ is the Kullback-Leibler divergence between instance $\inst{j}$ and instance $\inst{i}$.
	By the well-known decomposition theorem of the divergence, we know that:
	\[
	\mathcal{K}(\inst{j},\inst{i})=\mathbb{E}^\inst{j}\left[\sum\limits_{t=1}^T\phivec_t[(h_1, a)]\right]\mathcal{K}(B_{1/2+\epsilon}, B_{1/2-\epsilon})\le 16\epsilon^2\mathbb{E}^\inst{j}\left[\sum\limits_{t=1}^T\phivec_t[(h_1, a)]\right],
	\]
	where $\mathcal{K}(B_{1/2+\epsilon}, B_{1/2-\epsilon})$ is the Kullback-Leibler divergence between two Bernoulli random variable with parameter $1/2+\epsilon$ and $1/2-\epsilon$.
	Now, we can upper bound $\mathbb{E}^\inst{j}\left[\sum_{t=1}^T\phivec_t[(h_1, a)]\right]$ in terms of the probability $\mathbb{P}^{\inst{j}}$ with the reverse Markov inequality, as follows:
	\begin{align*}
		\mathbb{E}^\inst{j}\left[\sum\limits_{t=1}^T\phivec_t[(h_1, a)]\right]&\le \mathbb{P}^{\inst{j}}\left[\sum\limits_{t=1}^T\phivec_t[(h_1, a)]\ge K\right](T-K)+K\\
		&\le \delta(T-K)+K.
	\end{align*}
	Thus, we can conclude that: 
	\begin{equation}\label{eq:lower_bound1}
		\mathbb{P}^\inst{i}\left[\sum\limits_{t=1}^T\phivec_t[(h_1, a)]\le K\right]\ge 1-\delta-2\epsilon\sqrt{2(\delta(T-K)+K)}.
	\end{equation}
	Now, we consider the receiver's regret in instance $\inst{i}$, which can be computed as:
	\[
	\Rr_T^{\inst{i}}=\epsilon\sum\limits_{t=1}^T\phivec_t[(h_0, b)].
	\]
	This, together with Equation~\eqref{eq:lower_bound1}, allows us to conclude that:
	\[
	\mathbb{P}^\inst{i}\left[\Rr_T^{\inst{i}}\ge \epsilon(T-K)\right]\ge 1-\delta-2\epsilon\sqrt{2(\delta(T-K)+K)}.
	\]
	By setting $K=\frac{T^{\alpha}}{8}$ and $\epsilon=\frac{T^{-\alpha/2}}{8}$, we can conclude that if
	\[
	\mathbb{P}^\inst{j}\left[\sum\limits_{t=1}^T\phivec_t[(h_0, a)]\le \frac{T^{\alpha}}{8}\right]\ge 1-\delta,
	\]
	then
	\[
	\mathbb{P}^\inst{i}\left[\Rr_T^{\inst{i}}\ge \frac{T^{1-\alpha/2}}{16}\right]\ge 1-\frac{\sqrt{2}}{16}-\delta\ge 0.91-\delta,
	\]
	where we used that $\frac{T^{1-\alpha/2}}{8}-\frac{T^{\alpha/2}}{64}\ge \frac{T^{1-\alpha/2}}{16}$ for $T\ge 1$ and that we can assume $\delta\le \frac{T^{\alpha-1}}{4}$.
\end{proof}